\documentclass{article} 
\usepackage{iclr2025_conference,times}

\usepackage{amsmath,amsfonts,bm}

\def\eqref#1{equation~\ref{#1}}

\def\1{\bm{1}}

\DeclareMathAlphabet{\mathsfit}{\encodingdefault}{\sfdefault}{m}{sl}
\SetMathAlphabet{\mathsfit}{bold}{\encodingdefault}{\sfdefault}{bx}{n}

\iclrfinalcopy

\usepackage{hyperref}
\usepackage{url}
\usepackage[utf8]{inputenc} 
\usepackage[T1]{fontenc}    
\usepackage{hyperref}       
\usepackage{url}            
\usepackage{booktabs}       
\usepackage{amsfonts}       
\usepackage{nicefrac}       
\usepackage{microtype}      
\usepackage{xcolor}         
\usepackage{makecell}
\usepackage{graphicx}
\usepackage{soul}
\usepackage{amsmath, amsthm, amssymb}
\usepackage[most]{tcolorbox}
\usepackage{paralist}
\usepackage{caption}
\usepackage{bm}
\usepackage{subcaption}
\usepackage[american]{babel}
\usepackage{microtype}
\usepackage{latexsym}
\usepackage{booktabs}
\usepackage{multirow}
\usepackage{array}
\usepackage{natbib}
\usepackage{appendix}
\usepackage{algorithm}
\usepackage{algpseudocode} 
\usepackage{wrapfig}

\newcommand{\ys}[1]{\textcolor[rgb]{0,0,0}{#1}}

\newcommand{\highlight}[1]{{\color{magenta}{} #1}}

\definecolor{lightblue}{HTML}{328bb7}
\definecolor{lightred}{HTML}{d7907b}

\title{PiCO: Peer Review in LLMs based on Consistency Optimization}

\author{Kun-Peng Ning$^1$, Shuo Yang$^1$, Yu-Yang Liu$^{1,*}$, Jia-Yu Yao$^1$, Zhen-Hui Liu$^1$, \\ \textbf{Yong-Hong Tian$^{1,2}$}, \textbf{Yibing Song}, \textbf{Li Yuan$^{1,2,}$\thanks{Corresponding author.}} \\
$^1$School of Electrical and Computer Engineering, Peking University\\
$^2$Peng Cheng Laboratory \\
\texttt{\{ningkp,shuo\_yang,leon0425\}@stu.pku.edu.cn}, \\
\texttt{\{liuyuyang13,jiayu\_yao,yhtian,yuanli-ece\}@pku.edu.cn}, \\
\texttt{yibingsong.cv@gmail.com}
}

\begin{document}

\maketitle

\begin{abstract}

Existing large language models (LLMs) evaluation methods typically focus on testing the performance on some closed-environment and domain-specific benchmarks with human annotations. In this paper, we explore a novel \textbf{unsupervised evaluation direction}, utilizing \textit{peer-review} mechanisms to measure LLMs automatically without any human feedback.
In this setting, both open-source and closed-source LLMs lie in the same environment, capable of answering unlabeled questions and
evaluating each other, where each LLM’s response score is jointly determined by other anonymous ones. 
During this process, we found that those answers that are more recognized by other ``reviewers'' (models) usually come from LLMs with stronger abilities, while these models can also evaluate others' answers more accurately.  
We formalize it as a \textit{consistency assumption}, \textit{i.e.}, the ability and score of the model usually have consistency. 
We exploit this to optimize each model's confidence, thereby re-ranking the LLMs to be closer to human rankings.
We perform experiments on multiple datasets with standard rank-based metrics, validating the effectiveness of the proposed approach. Our code is released at \url{https://github.com/PKU-YuanGroup/PiCO}.


\end{abstract}

\section{Introduction}

\begin{quote}
    \centering
    Goodhart's Law: \emph{``When a measure becomes a target, it ceases to be a good measure.''}\\
\end{quote}

Large language models (LLMs) \citep{brown2020language,achiam2023gpt,bubeck2023sparks,touvron2023llama,ning2025gpt} have achieved remarkable success across a variety of real-world applications \citep{zhao2023survey,liu2023pre,ouyang2022training,yao2023llm,ning2024sparse,yang2024parameter}. With the increasingly widespread application of these models, there is an urgent need for an effective evaluation method to ensure that their performance and usability meet the growing demands. To assess the ability level of LLMs, a large number of evaluation benchmarks have been proposed by using
some small and domain-specific datasets with human-curated labels, such as MMLU \citep{hendrycks2020measuring}, HELM \citep{liang2022holistic}, Big-Bench \citep{srivastava2022beyond}, GLUE \citep{wang2018glue}. However, these benchmarks can only measure LLMs' core capability on a confined set of tasks (e.g. multi-choice knowledge or retrieval questions), which fails to assess their alignment with human preference in open-ended tasks adequately \citep{chiang2023vicuna,li2023prd,nakano2021webgpt}. 
On the other hand, these evaluations may suffer from \textit{benchmark leakage} issue, referring that the evaluation data is unknowingly used for model training, which can also lead to misleading evaluations \citep{wei2023skywork,zhou2023don}. Therefore, blindly improving scores on these public benchmarks cannot always yield a large language model that truly satisfies human requirements. 

For assessing human preferences, recent studies have focused on building crowdsourced battle platforms with human ratings as the primary evaluation metric. Typical platforms include Chatbot Arena \citep{zheng2023judging}, MT-Bench \citep{zheng2023judging}, and AlpacaEval \citep{li2023alpacaeval}. It constructs anonymous battles between chatbots in real-world scenarios, where users engage in conversations with two chatbots at the same time and rate their responses based on personal preferences. While human evaluation is the gold standard for measuring human preferences, it is exceptionally slow and costly \citep{zheng2023judging,ning2022active}. In addition, adding a new LLM to the crowdsourced battle platforms also poses a cold-start issue \citep{chang2023survey}. Thus, a fundamental question arises: \textit{can we construct an unsupervised LLMs evaluation system without relying on any human feedback}?

Actually, in real human evaluation systems, people build the human-ability hierarchy based on different empirical assumptions. For example, majority voting \citep{feldman2006majority,boyer1991mjrty,surowiecki2005wisdom} and rating voting \citep{allahbakhsh2012rating} methods are widely used during the decision-making process, which are based on the wisdom of the crowds \citep{surowiecki2005wisdom,budescu2015identifying,weller2007cultural} and have been proven to lead to better results than that of an individual. Moreover, in the established practice of \textit{peer-review} in academic research, scholars evaluate their academic level rankings based on the \textit{consistency assumption}, \textit{i.e.}, scholars with stronger abilities usually have stronger persuasiveness for evaluating others, and these scholars can also obtain higher achievements. This paper attempts to explore whether a similar phenomenon exists in the LLMs evaluation systems.  
\begin{figure*}[t]
    \centering
    \includegraphics[width=0.95\textwidth]{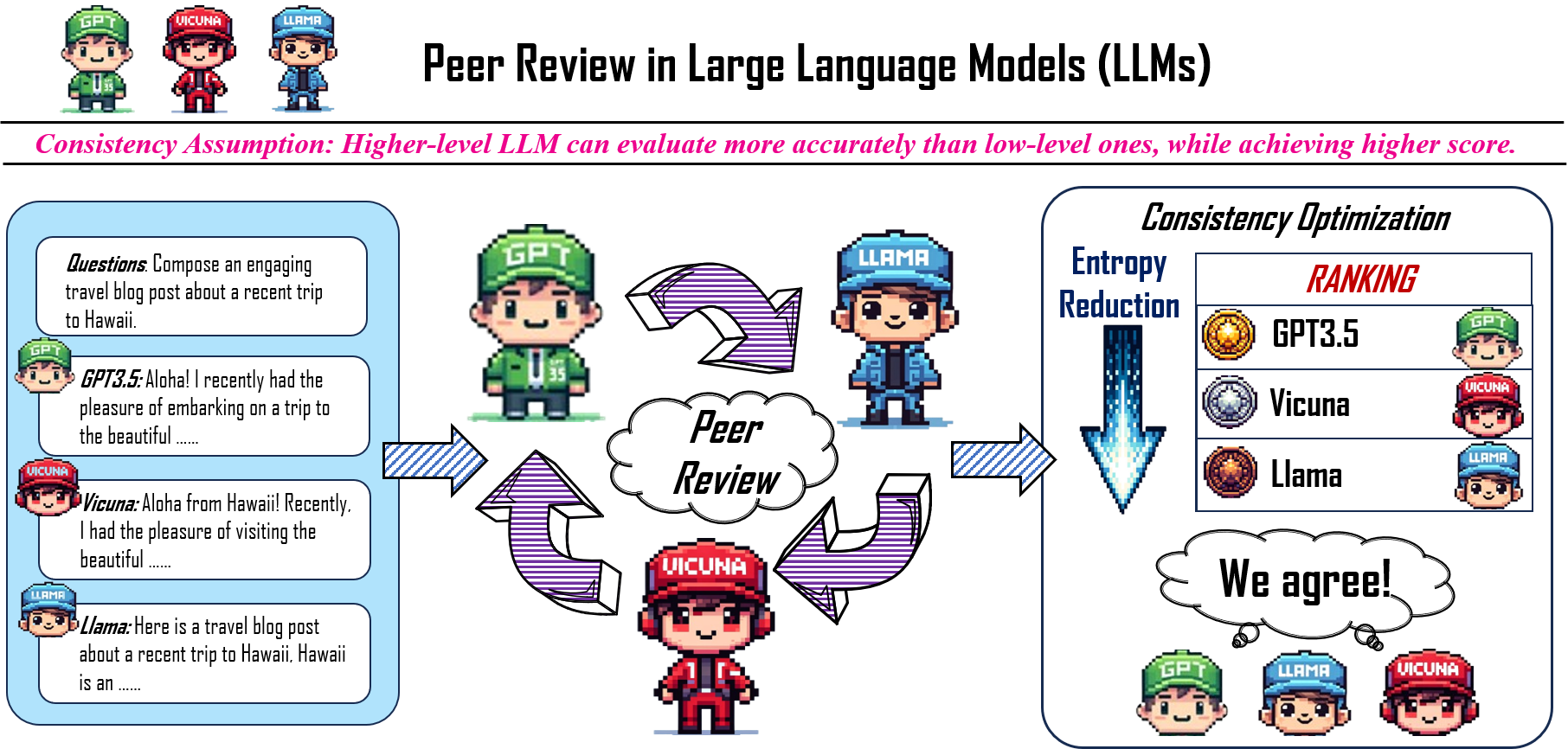}
    \caption{The framework of PiCO. In this framework, both open-source and closed-source LLMs lie in the same environment, capable of answering unlabeled questions and evaluating each other, where each LLM's response score is jointly determined by other anonymous ones. We assign each LLM a learnable capability weight to optimize the score ranking based on the \textit{consistency assumption}, while reducing the entropy of the \textit{peer-review} evaluation system. The goal is to find a final score ranking that all LLMs ``agree'' it.}
    \label{fig:peer-review}
    \vspace{-0.35cm}
\end{figure*}

In this paper, we propose \textbf{PiCO}, a \textbf{P}eer review approach \textbf{i}n LLMs based on \textbf{C}onsistency \textbf{O}ptimization.
In this setting, LLMs themselves act as ``reviewers'', engaging in mutual assessments to achieve comprehensive, efficient, and performance evaluations without relying on manually annotated data. 
This method aims to address the limitations of existing evaluation approaches and provide insights into LLMs' real-world capabilities.
As shown in Figure \ref{fig:peer-review}, both open-source and closed-source LLMs lie in the same environment and answer the open-ended questions from an unlabeled dataset.
Then, we construct anonymous answer pairs, while randomly selecting other LLMs as ``reviewers'' to evaluate both responses with a learnable confidence weight $w$. Finally, we employ this weight and calculate the response scores $G$ for each LLM based on the weighted joint evaluation. 
It is worth noting that the whole \textit{peer-review} process works in an unsupervised way, and our goal is to optimize the confidence weights $w$ that re-rank the LLMs to be closer to human rankings.

To achieve this, we formalize it as a constrained optimization based on the consistency assumption. We maximize the consistency of each LLM's capability $w$ and score $G$ while adjusting the final ranking to align with human preference more closely. \highlight{The key assumption behind this is that high-level LLM can evaluate others' answers more accurately (confidence) than low-level ones, while higher-level LLM can also achieve higher answer-ranking scores.} As a result, the entropy (controversy) of the whole \textit{peer-review} evaluation system can be minimized. In other words, the consistency optimization aims to find a final score ranking that all LLMs have no ``disputes'' regarding.

We perform experiments on multiple crowdsourcing datasets with standard rank-based metrics, the results demonstrate that the proposed PiCO framework can effectively obtain a large language models' leaderboard closer to human preferences. The contributions of this paper can be summarized as follows:



\begin{itemize}
    \setlength{\itemsep}{0pt} 
    \setlength{\parsep}{0pt}  
    \item We explore a novel unsupervised LLM evaluation direction without human feedback, \textit{i.e.}, utilizing \textit{peer-review} mechanisms to measure LLMs automatically. All LLMs can answer unlabeled questions and evaluate each other.
    \item A constrained optimization based on the consistency assumption is proposed to re-rank the LLMs to be closer to human rankings.
    \item We conduct extensive experiments on three crowdsourcing datasets with three standard rank-based metrics validating the effectiveness of the proposed PiCO approach.
\end{itemize}



\begin{figure*}[t]
    \centering
    \includegraphics[width=1\textwidth]{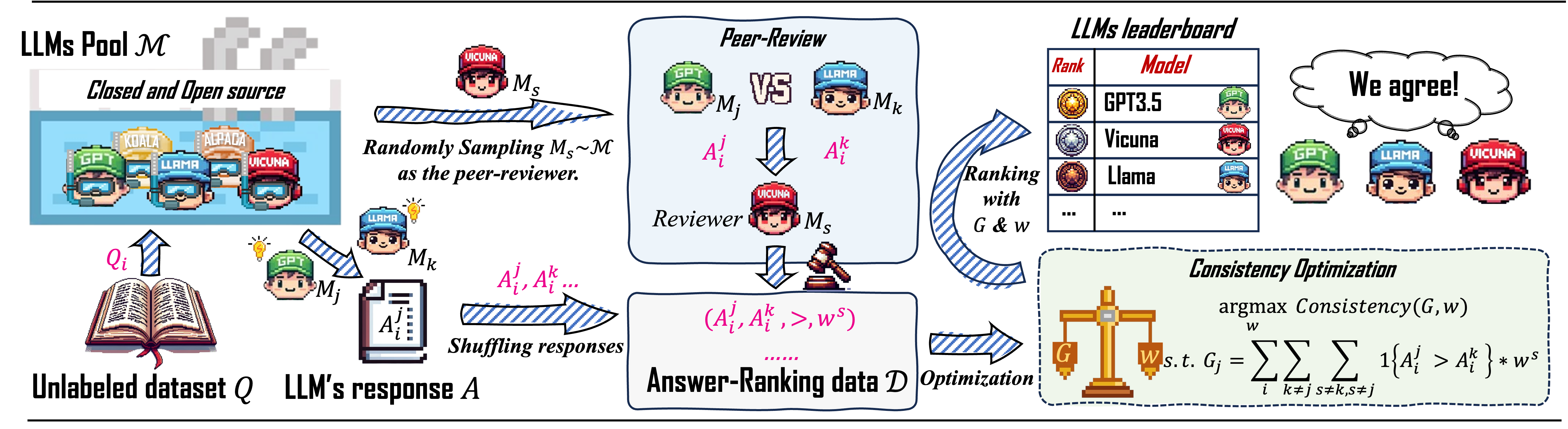}
    \caption{The pipeline of the PiCO. It is mainly composed of two components: the peer-review and consistency optimization stages. Specifically, in the peer-review stage, the unlabeled dataset $\mathcal{Q}$ and the LLMs pool $\mathcal{M}$ are given. Then, we let all LLMs answer each unlabeled question to obtain the response set $\mathcal{A}$. We shuffle the set and construct anonymous answer pairs, while randomly selecting other LLMs to evaluate both responses with a learnable confidence $w$. As a result, we can obtain the answer-ranking data $\mathcal{D}$ which is a quadruple that records the partial order between two answers and the evaluator's confidence weight. In the consistency optimization stage, we update the parameter $w$ by maximizing the consistency of each LLM's capability and score, while re-ranking the LLMs to be closer to human rankings. }
    \label{fig:peer-review-framwork}
\end{figure*}

\section{The Proposed Approach}
\subsection{Problem Definition}


This paper aims to re-rank the ability of LLMs to be closer to human (ground-truth) rankings $\mathcal{R}^*$ in an unsupervised way (without relying on any human annotations). Specifically, we have a large language models (LLMs) pool $\mathcal{M}=\{M_j\}_{j=1}^m$, which includes both open-source and closed-source models. Write ${M}_1 \succ {M}_2$ to indicate that the LLM ${M}_1$ has stronger capabilities than the LLM ${M}_2$. 
Thus, we can assume that the ground-truth ranking $\mathcal{R}^*$ is as follows,
\begin{equation}
\label{eq:human_ranking}
    \mathcal{R}^* := [{M}_1 \succ {M}_2 \succ {M}_3 \succ ... \succ {M}_m].
\end{equation}
Assuming that the learned ranking $\hat{\mathcal{R}}$ by different evaluation methods is as follows,
\begin{equation}
\label{eq:learned_ranking}
    \hat{\mathcal{R}} := [{M}_3 \succ {M}_1 \succ {M}_2 \succ ... \succ {M}_m].
\end{equation}
The goal is to learn an LLM ranking $\hat{\mathcal{R}}$ that aligns with human ranking $\mathcal{R}^*$ as much as possible. 

\subsection{Algorithm Details}
The pipeline of the proposed PiCO, depicted in Figure \ref{fig:peer-review-framwork}, involves peer-review and consistency optimization stages. Next, we will introduce the two stages in detail. 

\textbf{Peer Review Stage.} 
In our \textit{peer-review} system, we consider an unsupervised LLM evaluation scenario with an unlabeled dataset $\mathcal{Q}$ consisting of $n$ open-ended questions, where $\mathcal{Q}=\{Q_i\}_{i=1}^n$. All LLMs will answer each unlabeled question to obtain the set $\mathcal{A}=\{\{A_i^j\}_{i=1}^n\}_{j=1}^m$, where $A_i^j$ is as follows, 
\begin{equation}
    A_i^j = M_j(Q_i)
\end{equation}
which infers the model $M_j$ response an answer $A_i^j$ with question $Q_i$. In addition, LLMs themselves also act as ``reviewers'' to evaluate other answers. Specifically, for the same question $Q_i \in \mathcal{Q}$, we randomly construct a battle pair $<A_i^j, A_i^k>$ for review. Each battle pair will randomly assign ``reviewers'' to determine the winners or declare ties, 
\begin{equation}
\label{eq:4}
    (A_i^k, A_i^s, >, w^j) =M_j(A_i^k;A_i^s|Q_i).
\end{equation}
Under the same question $Q_i$, the quadruples $(A_i^k, A_i^s, >, w^j)$ indicate that the ``reviewer'' $M_j$ believes that the answer $A_i^k$ is better than answer $A_i^k$ with a confidence $w^j$. Thus, we can collect the answer-ranking data $\mathcal{D}$ as follows, 
\begin{equation}
\label{eq:D}
    \mathcal{D}=\left\{({A}_i^k,{A}_i^s,>,w^j)\right\}_{i\sim\mathcal{Q},j,k,M_j\sim\mathcal{M}},
\end{equation}
where $i$ denotes the question index, and $j,k,s$ indicate the model indices. $w^s \in (0,1]$ is a learnable confidence weight of model $M_s$, and $>$ is a partial order relationship from $\{>, <, =\}$. After that, we can calculate the response score $G_j$ of each LLM, 
\begin{equation}
\label{eq:6}
    G_j = \sum_{(A_i^k, A_i^s, >, w^j) \sim \mathcal{D}} \mathbf{1}\{A_i^j > A_i^k\} \cdot w^s,
\end{equation}
where $\mathbf{1}\{\cdot\}$ is the indicator function that the value is 1 when the condition is met, otherwise, it is 0. We can define the LLM $M_1$ is better than $M_2$ as its score is larger, \textit{i.e.}, $M_1 \succ M_2 := G_1 > G_2$. Thus, we can re-write the learned LLM ranking $\hat{\mathcal{R}}$ as follows, 
\begin{equation}
    \hat{\mathcal{R}} := [{G}_3 > {G}_1 > {G}_2 > ... > {G}_m].
\end{equation}
Thus, the goal is to learn the confidence weights $w$ to adjust the final ranking $\hat{\mathcal{R}}$ to be closer to ground-truth ranking $\mathcal{R}^*$.

\begin{table*}[!t]
    \renewcommand{\arraystretch}{1}
    \setlength{\tabcolsep}{1pt}
    \centering
    \caption{Validation of consistency assumption. Performance comparison of Backward, Uniform, Forward weight voting, and Consistency Optimization methods with two metrics across three datasets.}
    \vspace{-0.2cm}
    \label{table: upper}
    \begin{tabular}{c | c c | c c | c c }
    \toprule
        \multirow{2}{*}{Methods} & \multicolumn{2}{c}{MT-Bench} & \multicolumn{2}{|c}{Chatbot Arena} & \multicolumn{2}{|c}{AlpacaEval}  \\ 
        ~ & $S(\uparrow)$ & $\tau(\uparrow)$ & $S(\uparrow)$ & $\tau(\uparrow)$ & $S(\uparrow)$ & $\tau(\uparrow)$ \\
        \hline
        Backward Weight & 0.70 & 0.50 & 0.72 & 0.52 & 0.69 & 0.50 \\
        Uniform Weight & 0.74 & 0.54 & 0.80 & 0.58 & 0.77 & 0.58 \\
        Forward Weight & 0.75 & 0.56 & 0.82 & 0.59 & 0.79 & 0.60 \\
        \textbf{Random Weight + Consistency Optimization} & \textbf{0.90} & \textbf{0.77} & \textbf{0.89} & \textbf{0.72} & \textbf{0.84} & \textbf{0.68} \\
        \toprule
    \end{tabular}
    \vspace{-0.35cm}
\end{table*}

\textbf{Validation of Consistency Assumption.} First of all, we start with a toy experiment to study the role of confidence $w$ in Table \ref{table: upper}. 
Specifically, we manually construct three methods: Backward Weight, Uniform Weight, and Forward Weight. That is, the ability weights of the model are respectively weighted forward ($w=[1,0.9,...,0]$), uniformly ($w=[1,1,...,1]$), and backward ($w=[0,0.1,...,1]$) according to the ground-truth human ranking. In other words, the Forward Weight means manually assigning higher weights to those models with stronger abilities, and so on for others. Then, we can calculate the response score $G_j$ for each model using Eq.\ref{eq:6}, and obtain the LLM ranking $\hat{\mathcal{R}}$. We measure the alignment between $\hat{\mathcal{R}}$ and $\mathcal{R}^*$ with Spearman's $S(\uparrow)$ and Kendall's $\tau(\uparrow)$ rank correlation coefficient in Table \ref{table: upper}. Note that this is an ideal experiment, as we only use the ground-truth human ranking to validate the feasibility of our idea. 

As shown in Table \ref{table: upper}, it can be observed that the Forward Weight achieves better results than the Uniform and Backward ones in all cases, while the Backward one always achieves worse results. \textbf{It validates that assigning larger weights to those models with stronger capabilities can obtain better results.} In other words, those answers that are more recognized by other ``reviewers'' (models) usually come from LLs with stronger abilities. We formalize it as a \textit{consistency assumption}, \textit{i.e.}, high-level LLM can evaluate others' answers more accurately (confidence) than low-level ones, while higher-level LLM can also achieve higher answer-ranking scores, the ability and score of the model usually have consistency. 

\textbf{Consistency Optimization Stage.} Based on this observation, we propose to maximize the consistency of each LLM's capability $w$ and score $G$ with constrained optimization as follows, 
\begin{align}
\label{eq:G}
&\underset{w}{\mathrm{argmax}}\,\, \text{Consistency}(G, w) \\
\text{s.t. }G_j &= \sum_{(A_i^j, A_i^k, >, w^s) \sim \mathcal{D}} \mathbf{1}\{A_i^j > A_i^k\} \cdot w^s \nonumber,
\end{align}
where the Pearson correlation \citep{sedgwick2012pearson} is used to measure the consistency between $w$ and $G$. Note that we only introduce this straightforward implementation to validate our idea of PiCO. Other more advanced strategies may be employed to further improve the performance.

\textbf{Discussion: } It is worth noting that the whole process (Eq. \ref{eq:D} and \ref{eq:G}) works in an unsupervised way. The only thing we do is to adaptively adjust the score of each LLM that match its abilities. Most importantly, we also validate the effectiveness of the proposed \textit{consistency optimization} in Table \ref{table: upper}. Specifically, we randomly initialize the ability weights and employ our \textit{consistency optimization} to adjust the weight. It can be observed that the learned $w$ by our consistency optimization algorithm (Eq.\ref{eq:G}) can further improve the performance of the evaluation system, making the LLM ranking $\hat{\mathcal{R}}$ closer to human ranking $\mathcal{R}^*$. 
Another intuitive example is as follows: in a real peer-review system, if the academic level of three scholars $a$, $b$, and $c$ satisfies the following relationship, $w^a>w^b>w^c$. So, in the ultimate ideal scenario, the ranking of the scores submitted by these three scholars should also be, $G_a>G_b>G_c$. In other words, the sorting of $G$ and $w$ satisfies high consistency. On the other hand, scholars with stronger abilities (\textit{i.e.}, scholar $a$) evaluate $A^b>A^c$ have stronger persuasiveness, so scholar $b$ should also receive higher weighted scores $1*w^a$.

\textbf{Reviewer Elimination Mechanism.}  
Realizing that not all LLMs have sufficient ability to evaluate the responses of other models. We thus introduce an unsupervised elimination mechanism to remove those LLMs that have low scores. It iteratively removes the lowest-scoring LLM from the ``reviewer queue'' for the next consistency optimization stage, until 60\% of models are eliminated. The discussion of the elimination mechanism can also be found in the Experiment \ref{section:3.4}. 

\section{Experiments}
\label{Experiments}

\textbf{Datasets.}
To validate the effectiveness of the proposed approach, we perform experiments on Chatbot Arena \citep{zheng2023judging}, MT-Bench \citep{zheng2023judging}, and AlpacaEval \citep{li2023alpacaeval}. The MT-Bench dataset assesses six LLMs' responses to 80 multi-category questions. The Chatbot Arena Conversations Dataset, with 33K conversations from 13K IPs during April-June 2023, evaluates real dialogue performance. AlpacaEval dataset integrates 805 evaluations from diverse tests (e.g., Self-Instruct \citep{wang2022self}, OASST, Anthropic’s helpful \citep{bai2022training}, Vicuna \citep{chiang2023vicuna} and Koala \citep{geng2023koala} test sets) to align evaluations real-world interactions \citep{dubois2023alpacafarm}.
These datasets are collected by crowdsourcing platforms from human feedback, so they have a ground-truth ranking LLMs $\mathcal{R}^*$ to measure the alignment performance of different evaluation methods.

\textbf{LLMs Pool.} In our experiments, we employ 15 LLMs with diverse architectures to construct the LLMs pool, including GPT-3.5-Turbo \citep{OpenAI2022}, WizardLM-13B \citep{xu2023wizardlm}, Guanaco-33B \citep{guanaco2023model}, Vicuna-7B \citep{chiang2023vicuna}, Vicuna-13B \citep{chiang2023vicuna}, Koala-13B \citep{koala13b2023}, Mpt-7B \citep{MosaicML2023Introducing}, gpt4all-13B \citep{gpt4all}, ChatGLM-6B \citep{zeng2022glm}, Oasst-sft-4-pythia-12B \citep{oasst_sft_4_pythia_12b}, FastChat-T5-3B \citep{zheng2023judging}, StableLM-7B \citep{stablelm_tuned_alpha_7b_2023}, Dolly-12B \citep{DatabricksBlog2023DollyV2}, LLaMA-13B \citep{touvron2023llama}, Alpaca-13B \citep{alpaca}. 
All models use the same prompt template, which can be found in Appendix \ref{appendix:B}.







\begin{table*}[!t]
    \renewcommand{\arraystretch}{1}
    \setlength{\tabcolsep}{1pt}
    \centering
    \caption{Comparison of all methods on three datasets under data volumes of 1, 0.7, and 0.4, where the top value is highlighted by bold font. Higher $S$ and $\tau$ scores indicate better performance, while a lower $H$ score signifies improved performance.}
    \vspace{-0.2cm}
    \label{table: 1, 0.7，0.4}
    \resizebox{\textwidth}{!}{\begin{tabular}{c| c c c| c c c| c c c}
        \toprule
        Datasets & \multicolumn{3}{c}{Chatbot Arena} & \multicolumn{3}{c}{MT-Bench} & \multicolumn{3}{c}{AlpacaEval} \\
        Methods & 1 & 0.7 & 0.4 & 1 & 0.7 & 0.4 & 1 & 0.7 & 0.4\\
        \toprule
        
        & \multicolumn{9}{c}{Spearman's Rank Correlation Coefficient $S(\uparrow)$} \\
        \toprule
        Majority Voting & $0.76^{\pm0.00}$ & $0.75^{\pm0.01}$ & $0.73^{\pm0.03}$ & $0.73^{\pm0.00}$ & $0.77^{\pm0.01}$ & $0.75^{\pm0.01}$ & $0.80^{\pm0.00}$ & $0.79^{\pm0.01}$ & $0.78^{\pm0.01}$ \\
        Rating Voting & $0.74^{\pm0.00}$ & $0.72^{\pm0.02}$ & $0.71^{\pm0.02}$ & $0.80^{\pm0.00}$ & $0.78^{\pm0.02}$ & $0.74^{\pm0.03}$ & $0.77^{\pm0.00}$ & $0.77^{\pm0.01}$ & $0.78^{\pm0.01}$ \\
        \hline
        GPTScore(flan-t5-xxl) & $-0.09^{\pm0.00}$ & $-0.09^{\pm0.01}$ & $-0.12^{\pm0.02}$ & $0.05^{\pm0.00}$ & $0.01^{\pm0.07}$ & $0.04^{\pm0.09}$ & $0.34^{\pm0.00}$ & $0.34^{\pm0.00}$ & $0.34^{\pm0.01}$ \\  
        GPTScore(davinci-002) & $0.15^{\pm0.00}$ & $0.13^{\pm0.02}$ & $-0.02^{\pm0.14}$ & $0.52^{\pm0.00}$ & $0.42^{\pm0.05}$ & $0.45^{\pm0.05}$ & $0.76^{\pm0.00}$ & $0.77^{\pm0.07}$ & $0.75^{\pm0.06}$ \\
        PandaLM & $0.43^{\pm0.00}$ & $0.44^{\pm0.03}$ & $0.44^{\pm0.10}$ & $0.50^{\pm0.00}$ & $0.50^{\pm0.08}$ & $0.52^{\pm0.17}$ & $0.57^{\pm0.00}$ & $0.55^{\pm0.01}$ & $0.48^{\pm0.08}$ \\
        PRD & $0.84^{\pm0.00}$ & $0.84^{\pm0.00}$ & $0.82^{\pm0.03}$ & $0.86^{\pm0.00}$ & $0.84^{\pm0.03}$ & $0.81^{\pm0.03}$ & $0.81^{\pm0.00}$ & $0.81^{\pm0.01}$ & $0.81^{\pm0.02}$ \\
        PRE & $0.86^{\pm0.00}$ & $0.86^{\pm0.01}$ & $0.86^{\pm0.01}$ & $0.86^{\pm0.00}$ & $0.84^{\pm0.03}$ & $0.82^{\pm0.04}$ & $0.83^{\pm0.00}$ & $0.81^{\pm0.01}$ & $0.83^{\pm0.02}$ \\
        Claude-3 (API) & $0.90^{\pm0.01}$ & $0.88^{\pm0.03}$ & $0.87^{\pm0.04}$ & $0.85^{\pm0.06}$ & $0.82^{\pm0.08}$ & $0.80^{\pm0.07}$ & $0.79^{\pm0.03}$ & $0.78^{\pm0.02}$ & $0.75^{\pm0.04}$ \\
        \textbf{PiCO (Ours)} & $\bm{0.90^{\pm0.00}}$ & $\bm{0.89^{\pm0.01}}$ & $\bm{0.89^{\pm0.01}}$ & $\bm{0.89^{\pm0.01}}$ & $\bm{0.89^{\pm0.01}}$ & $\bm{0.84^{\pm0.11}}$ & $\bm{0.84^{\pm0.00}}$ & $\bm{0.83^{\pm0.03}}$ & $\bm{0.85^{\pm0.01}}$ \\ 
        \toprule
        & \multicolumn{9}{c}{Kendall's Rank Correlation Coefficient $\tau(\uparrow)$} \\
        \toprule
        Majority Voting & $0.58^{\pm 0.00}$ & $0.56^{\pm 0.02}$ & $0.52^{\pm 0.05}$ & $0.56^{\pm 0.00}$ & $0.61^{\pm 0.02}$ & $0.60^{\pm 0.02}$ & $0.62^{\pm 0.00}$ & $0.60^{\pm 0.02}$ & $0.58^{\pm 0.02}$ \\
        Rating Voting & $0.54^{\pm 0.00}$ & $0.53^{\pm 0.02}$ & $0.52^{\pm 0.02}$ & $0.58^{\pm 0.00}$ & $0.57^{\pm 0.02}$ & $0.54^{\pm 0.01}$ & $0.58^{\pm 0.00}$ & $0.57^{\pm 0.01}$ & $0.57^{\pm 0.02}$ \\
        \hline
        GPTScore(flan-t5-xxl) & $-0.06^{\pm 0.00}$ & $-0.06^{\pm 0.02}$ & $-0.09^{\pm 0.02}$ & $-0.05^{\pm 0.00}$ & $-0.07^{\pm 0.05}$ & $-0.02^{\pm 0.06}$ & $0.25^{\pm 0.00}$ & $0.26^{\pm 0.01}$ & $0.26^{\pm 0.01}$ \\
        GPTScore(davinci-002) & $0.20^{\pm 0.00}$ & $0.23^{\pm 0.02}$ & $0.03^{\pm 0.11}$ & $0.36^{\pm 0.00}$ & $0.30^{\pm 0.05}$ & $0.31^{\pm 0.05}$ & $0.60^{\pm 0.08}$ & $0.61^{\pm 0.05}$ & $0.59^{\pm 0.08}$ \\
        PandaLM & $0.30^{\pm 0.00}$ & $0.31^{\pm 0.03}$ & $0.31^{\pm 0.07}$ & $0.39^{\pm 0.00}$ & $0.37^{\pm 0.06}$ & $0.40^{\pm 0.12}$ & $0.41^{\pm 0.00}$ & $0.39^{\pm 0.02}$ & $0.32^{\pm 0.05}$ \\
        PRD & $0.68^{\pm 0.00}$ & $0.69^{\pm 0.01}$ & $0.67^{\pm 0.03}$ & $0.68^{\pm 0.06}$ & $0.66^{\pm 0.02}$ & $0.63^{\pm 0.03}$ & $0.64^{\pm 0.00}$ & $0.63^{\pm 0.03}$ & $0.63^{\pm 0.02}$ \\
        PRE & $0.71^{\pm 0.00}$ & $0.73^{\pm 0.02}$ & $0.72^{\pm 0.02}$ & $0.68^{\pm 0.00}$ & $0.68^{\pm 0.02}$ & $0.65^{\pm 0.03}$ & $0.64^{\pm 0.00}$ & $0.66^{\pm 0.01}$ & $0.66^{\pm 0.03}$ \\
        Claude-3 (API)         & $0.76^{\pm 0.04}$ & $0.72^{\pm 0.05}$ & $0.70^{\pm 0.07}$ & $0.67^{\pm 0.07}$ & $0.66^{\pm 0.11}$ & $0.61^{\pm 0.10}$ & $0.64^{\pm 0.06}$ & $0.61^{\pm 0.04}$ & $0.66^{\pm 0.06}$ \\
        \textbf{PiCO (Ours)} & \bm{$0.77^{\pm 0.00}$} & \bm{$0.76^{\pm 0.01}$} & \bm{$0.77^{\pm 0.02}$} & \bm{$0.72^{\pm 0.01}$} & \bm{$0.72^{\pm 0.03}$} & \bm{$0.70^{\pm 0.12}$} & \bm{$0.68^{\pm 0.00}$} & \bm{$0.66^{\pm 0.04}$} & \bm{$0.67^{\pm 0.02}$} \\
        \toprule
        
        & \multicolumn{9}{c}{Permutation Entropy $H(\downarrow)$}\\
        \toprule
        Majority Voting & $1.27^{\pm0.05}$ & $1.30^{\pm0.03}$ & $1.36^{\pm0.06}$ & $1.37^{\pm0.03}$ & $1.30^{\pm0.06}$ & $1.27^{\pm0.04}$ & $1.26^{\pm0.02}$ & $1.28^{\pm0.03}$ & $1.29^{\pm0.03}$ \\
        Rating Voting & $1.39^{\pm0.02}$ & $1.43^{\pm0.03}$ & $1.42^{\pm0.07}$ & $1.32^{\pm0.03}$ & $1.35^{\pm0.04}$ & $1.38^{\pm0.04}$ & $1.34^{\pm0.03}$ & $1.37^{\pm0.03}$ & $1.34^{\pm0.08}$ \\
        \hline
        GPTScore(flan-t5-xxl) & $1.68^{\pm0.01}$ & $1.68^{\pm0.02}$ & $1.65^{\pm0.02}$ & $1.72^{\pm0.02}$ & $1.70^{\pm0.02}$ & $1.68^{\pm0.03}$ & $1.55^{\pm0.02}$ & $1.57^{\pm0.03}$ & $1.60^{\pm0.01}$ \\
        GPTScore(davinci-002) & $1.54^{\pm0.02}$ & $1.64^{\pm0.02}$ & $1.68^{\pm0.05}$ & $1.51^{\pm0.02}$ & $1.61^{\pm0.01}$ & $1.61^{\pm0.04}$ & $1.25^{\pm0.02}$ & $1.23^{\pm0.08}$ & $1.26^{\pm0.14}$ \\
        PandaLM & $1.65^{\pm0.01}$ & $1.64^{\pm0.02}$ & $1.63^{\pm0.05}$ & $1.55^{\pm0.03}$ & $1.59^{\pm0.05}$ & $1.52^{\pm0.08}$ & $1.56^{\pm0.01}$ & $1.58^{\pm0.01}$ & $1.64^{\pm0.05}$ \\
        PRD & $1.15^{\pm0.04}$ & $1.12^{\pm0.05}$ & $1.13^{\pm0.06}$ & $1.15^{\pm0.05}$ & $1.17^{\pm0.06}$ & $1.23^{\pm0.04}$ & $1.21^{\pm0.04}$ & $1.22^{\pm0.06}$ & $1.23^{\pm0.07}$ \\ 
        PRE & $1.07^{\pm0.01}$ & $1.03^{\pm0.03}$ & $1.06^{\pm0.04}$ & $1.17^{\pm0.04}$ & $1.13^{\pm0.05}$ & $1.19^{\pm0.05}$ & $1.18^{\pm0.03}$ & $1.21^{\pm0.04}$ & $1.15^{\pm0.05}$ \\
        \textbf{PiCO (Ours)} & $\bm{0.94^{\pm0.02}}$ & $\bm{0.96^{\pm0.04}}$ & $\bm{0.95^{\pm0.08}}$ & $\bm{1.01^{\pm0.07}}$ & $\bm{1.02^{\pm0.11}}$ & $\bm{1.06^{\pm0.24}}$ & $\bm{1.17^{\pm0.02}}$ & $\bm{1.17^{\pm0.08}}$ & $\bm{1.13^{\pm0.05}}$ \\
        \bottomrule
    \end{tabular}}
\end{table*}

\textbf{Baselines.} To validate the effectiveness of the proposed PiCO approach, we compare the following methods in the experiments. 
\vspace*{-0.5\baselineskip}
\begin{itemize}
    \setlength\itemsep{0em}
    \setlength{\parsep}{0pt}  
    \item \textit{The wisdom of the crowds}: The two methods that perform LLMs evaluation based on the wisdom of the crowds \citep{surowiecki2005wisdom,budescu2015identifying,weller2007cultural} are compared in this experiment. \textit{1)} \textbf{Majority Voting} \citep{surowiecki2005wisdom}: Multiple review models vote for the better answer for the same response pair, and the model with the most votes gets 1 score; \textit{2)}\textbf{ Rating Voting} \citep{allahbakhsh2012rating}: Multiple review models also vote on the same response pair, and the number of votes obtained is the score. 
    
    \item \textit{State-of-the-art methods}: The four recent SOTA methods of using either single or multiple models for self-evaluation are compared in this experiment. 
    \textbf{PandaLM} \citep{wang2023pandalm}: It is a fine-tuned language model based on Llama-7b designed for the preference judgment tasks to evaluate and optimize LLMs. 
    \textbf{GPTScore} \citep{fu2023gptscore}: It employs generative pre-trained models to assess the quality of generated text. It calculates the likelihood that the text was generated in response to specific instructions and context, indicative of high quality. In our implementation, GPT-3 (davinci-002) and flan-t5-xxl serve as the base models. 
    \textbf{PRD} \citep{li2023prd}: It transforms the LLMs win rates into weights for competitive ranking, while evaluating each LLM based on its preference for all possible pairs of answers, enabling a tournament-style ranking system.
    \textbf{PRE} \citep{chu2024pre}: It employs a supervised process to evaluate LLMs using a qualification exam, aggregates their scores based on accuracy, and assigns weights accordingly. 
    \textbf{Claude-3 (API):} Another SOTA closed-source LLM developed by Anthropic. 
    \textbf{PiCO (Ours)}: the proposed approach in this paper. 
\end{itemize}
\vspace*{-0.5\baselineskip}

\textbf{Metrics.} For all experiments, we employ three popular rank-based metrics to evaluate the aforementioned experimental setups and our PiCO method: \textbf{Spearman's Rank Correlation Coefficient $S(\uparrow)$} \citep{lehman2013jmp}, \textbf{Kendall's Rank Correlation Coefficient $\tau(\uparrow)$} \citep{kendall1938new} and \textbf{Permutation Entropy $H(\downarrow)$} \citep{bandt2002permutation}. The details of these metrics can be found in the Appendix \ref{appendix:Metric}. Moreover, we perform the experiments for 4 runs and record the average results over 4 seeds ($seed=1,2,3,4$).

\subsection{Performance Comparison}
We validate the effectiveness of the proposed PiCO method on three datasets by comparing the following two types of methods, \textit{i.e.}, the wisdom of the crowds and recent SOTA LLMs evaluation methods. The average results with different rank-based metrics and datasets are demonstrated in Table \ref{table: 1, 0.7，0.4}. The ratios of response sets $\mathcal{D}$ are 1, 0.7, and 0.4, respectively.

The results presented in Table \ref{table: 1, 0.7，0.4} demonstrate that the proposed PiCO method consistently outperforms competing approaches across most evaluated metrics, including surpassing all baselines, such as \textbf{Claude-3 (API)}. Specifically, PiCO achieves improvements of 0.027, 0.047, and 0.14 on Spearman's Rank Correlation Coefficient, Kendall's Rank Correlation Coefficient, and Permutation Entropy metrics, respectively, compared to the runner-up. These results underscore the superiority of aggregating evaluations from multiple models, such as Majority Voting, Rating Voting, PRD, and PRE, as opposed to relying solely on single-model methods like GPTScore and PandaLM. 
This collective model approach, leveraging 'the wisdom of the crowds', aligns with human rankings more accurately in our open-question evaluation framework.

In comparison with existing SOTA evaluation methods(\emph{i.e.,} PRD and PRE), it is evident that PiCO exhibits improvements across various evaluation metrics. Despite PRD's adjustment of model weights based on their win rates and PRE's reliance on supervised human feedback data to assign weights through a qualification exam, neither method achieves performance superior to the fully unsupervised PiCO approach. These methods rely on predefined criteria and human feedback, potentially leading to biases or suboptimal performance. In contrast, PiCO leverages unsupervised learning techniques, allowing it to autonomously adapt and discover patterns in the data without explicit human intervention.


It is important to highlight that PandaLM, a language model equipped with 7 billion parameters, was fine-tuned using labels generated by GPT-3.5-turbo as the ground truth, achieving stable performance across various datasets. However, in our unsupervised, open-ended experimental setup, which focuses on ranking-based metrics, GPTScore exhibits less robustness regardless of whether the base model is GPT-3 (davinci-002) or flan-t5-xx.

\begin{figure*}[!t]
	\centering
	\begin{subfigure}{0.325\linewidth}
		\centering
		\includegraphics[width=0.9\linewidth]{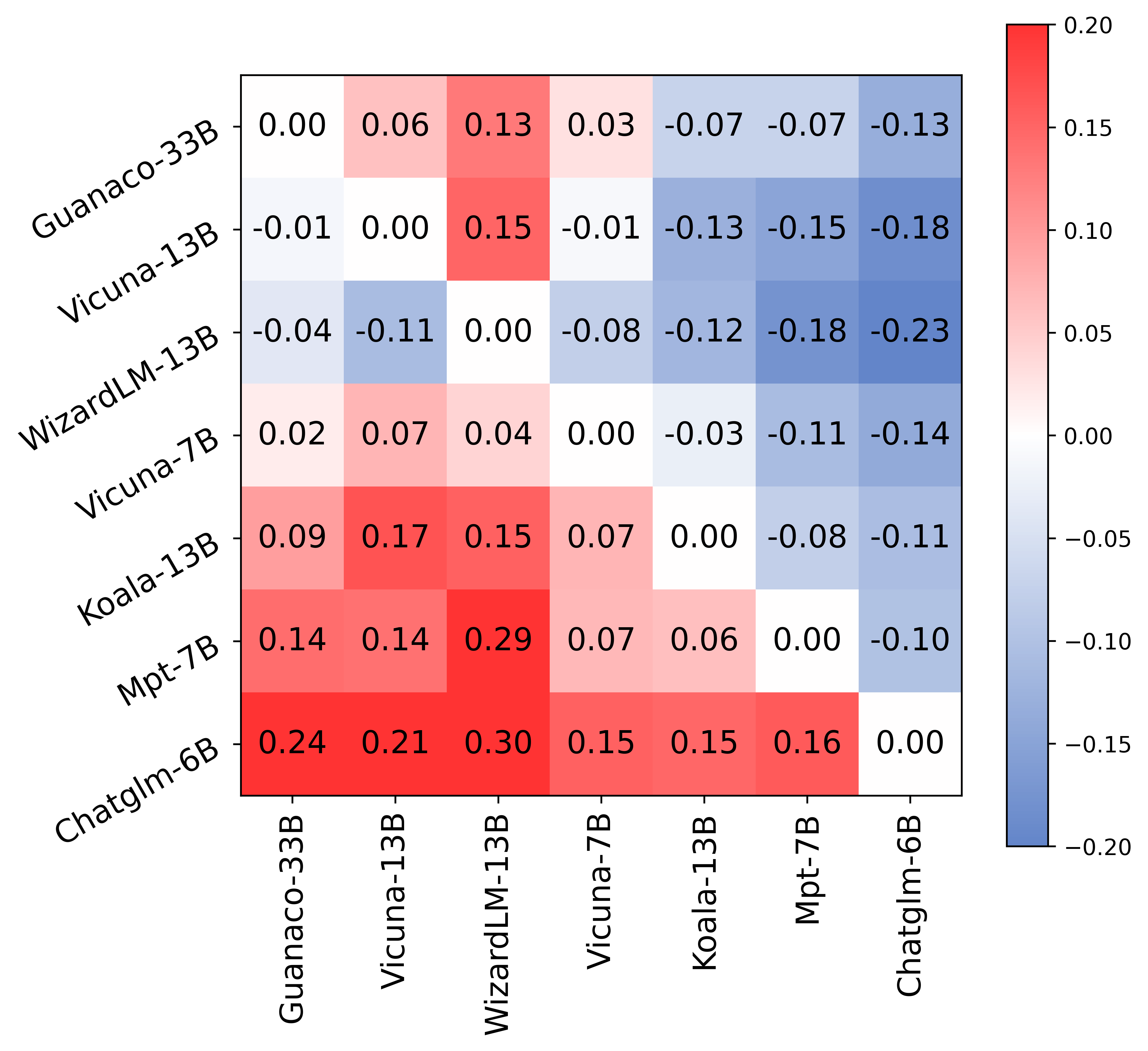}
		\caption{ChatBot Arena (PG)}
		\label{chutian2} 
	\end{subfigure}%
	\begin{subfigure}{0.325\linewidth}
		\centering
		\includegraphics[width=0.9\linewidth]{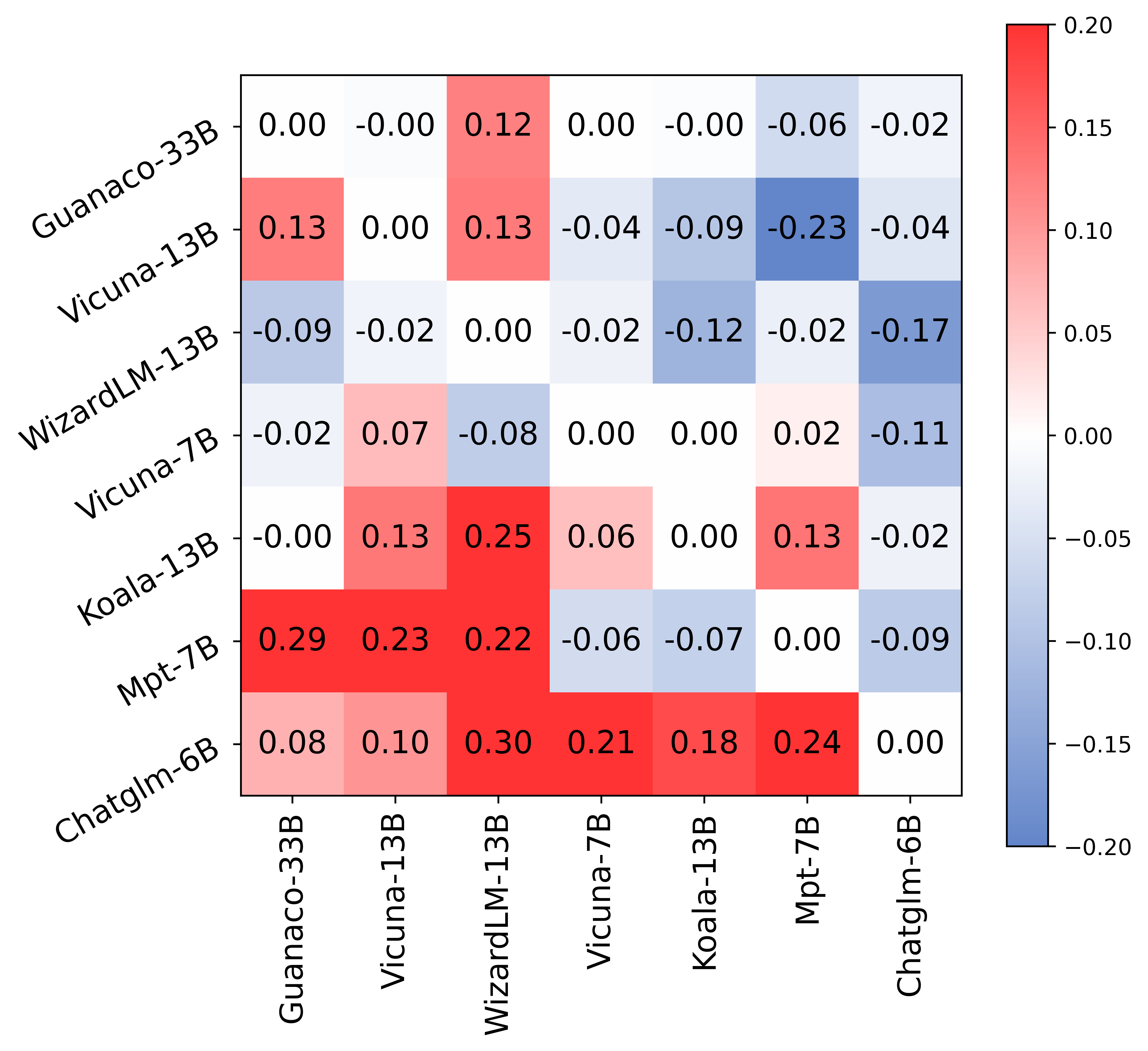}
		\caption{MT-Bench (PG)}
		\label{chutian1} 
	\end{subfigure}%
	\begin{subfigure}{0.325\linewidth}
		\centering
		\includegraphics[width=0.9\linewidth]{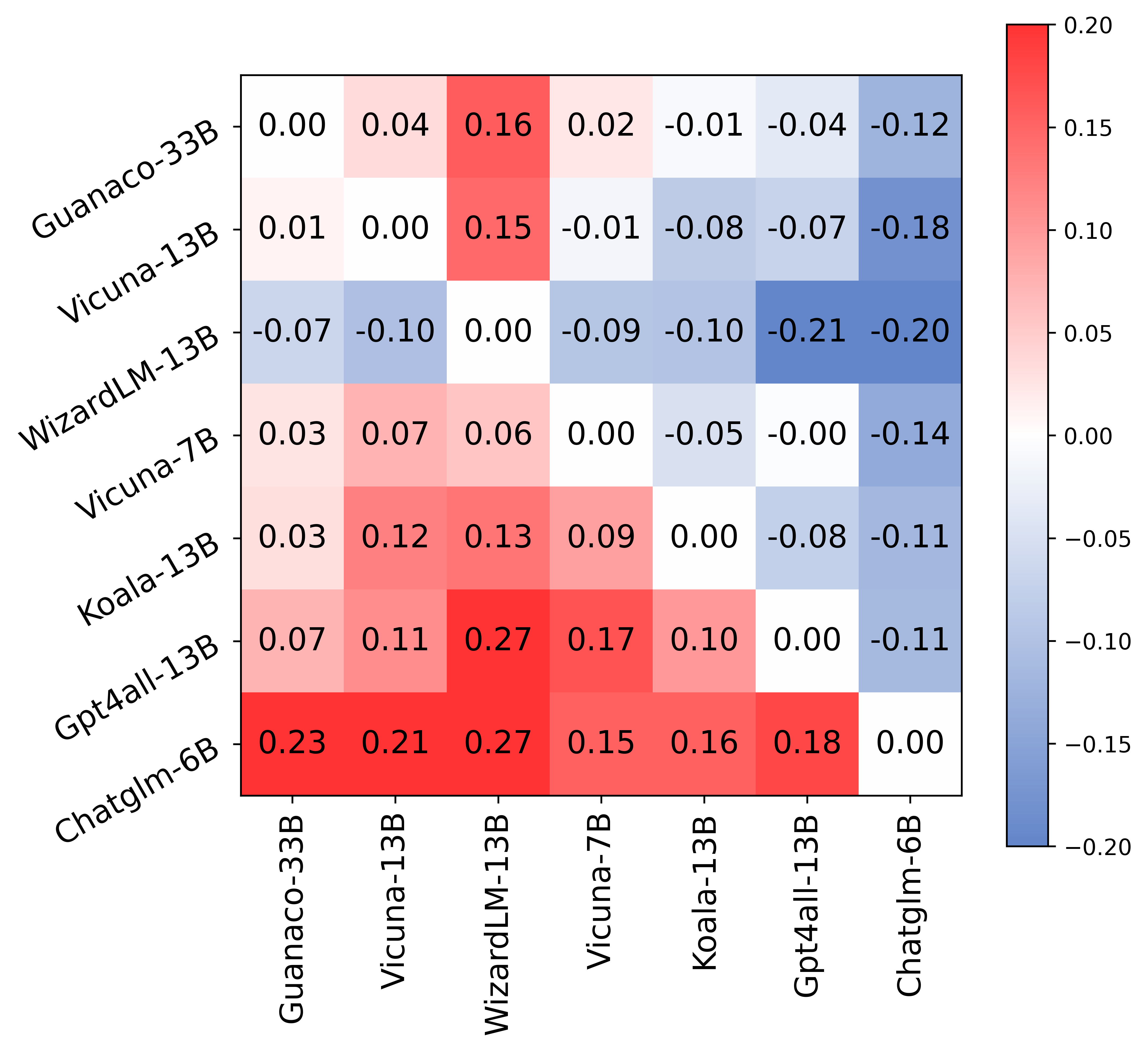}
		\caption{AlpacaEval (PG)}
		\label{chutian3} 
	\end{subfigure}

	\begin{subfigure}{0.325\linewidth}
		\centering
		\includegraphics[width=0.9\linewidth]{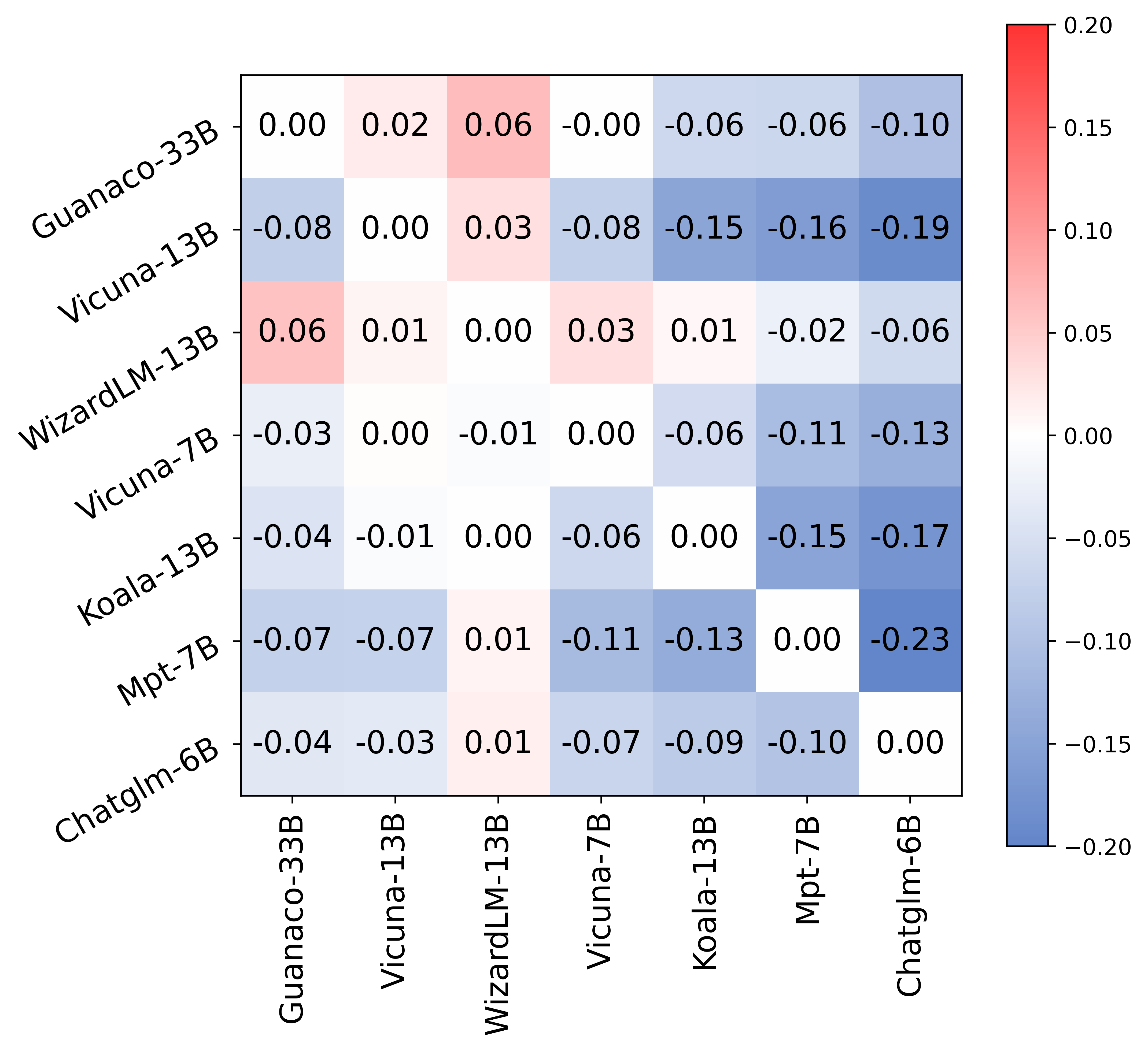}
		\caption{ChatBot Arena (weighted PG)}
		\label{chutian5} 
	\end{subfigure}%
	\begin{subfigure}{0.325\linewidth}
		\centering
		\includegraphics[width=0.9\linewidth]{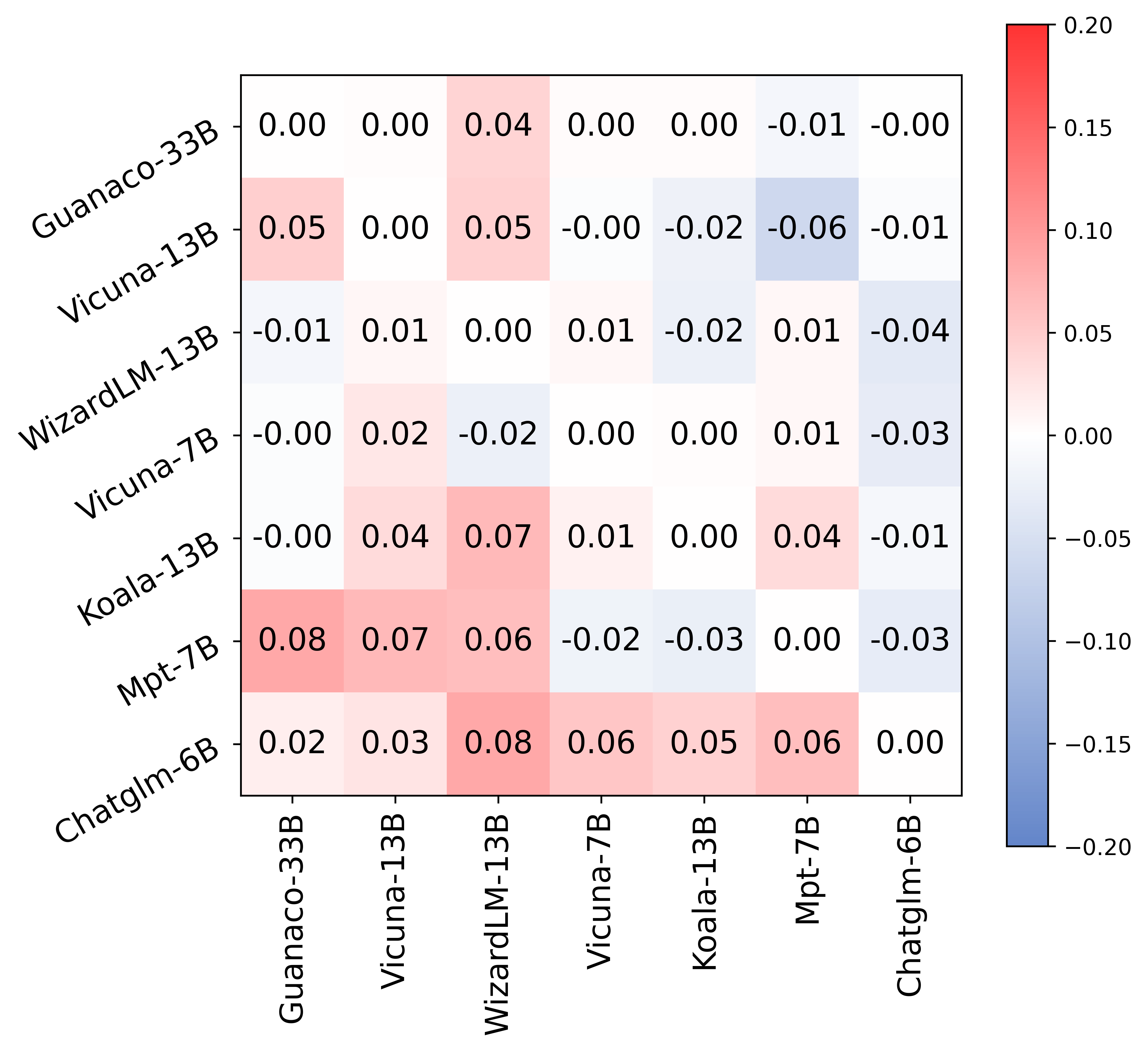}
		\caption{MT-Bench (weighted PG)}
		\label{chutian4} 
	\end{subfigure}%
	\begin{subfigure}{0.325\linewidth}
		\centering
		\includegraphics[width=0.9\linewidth]{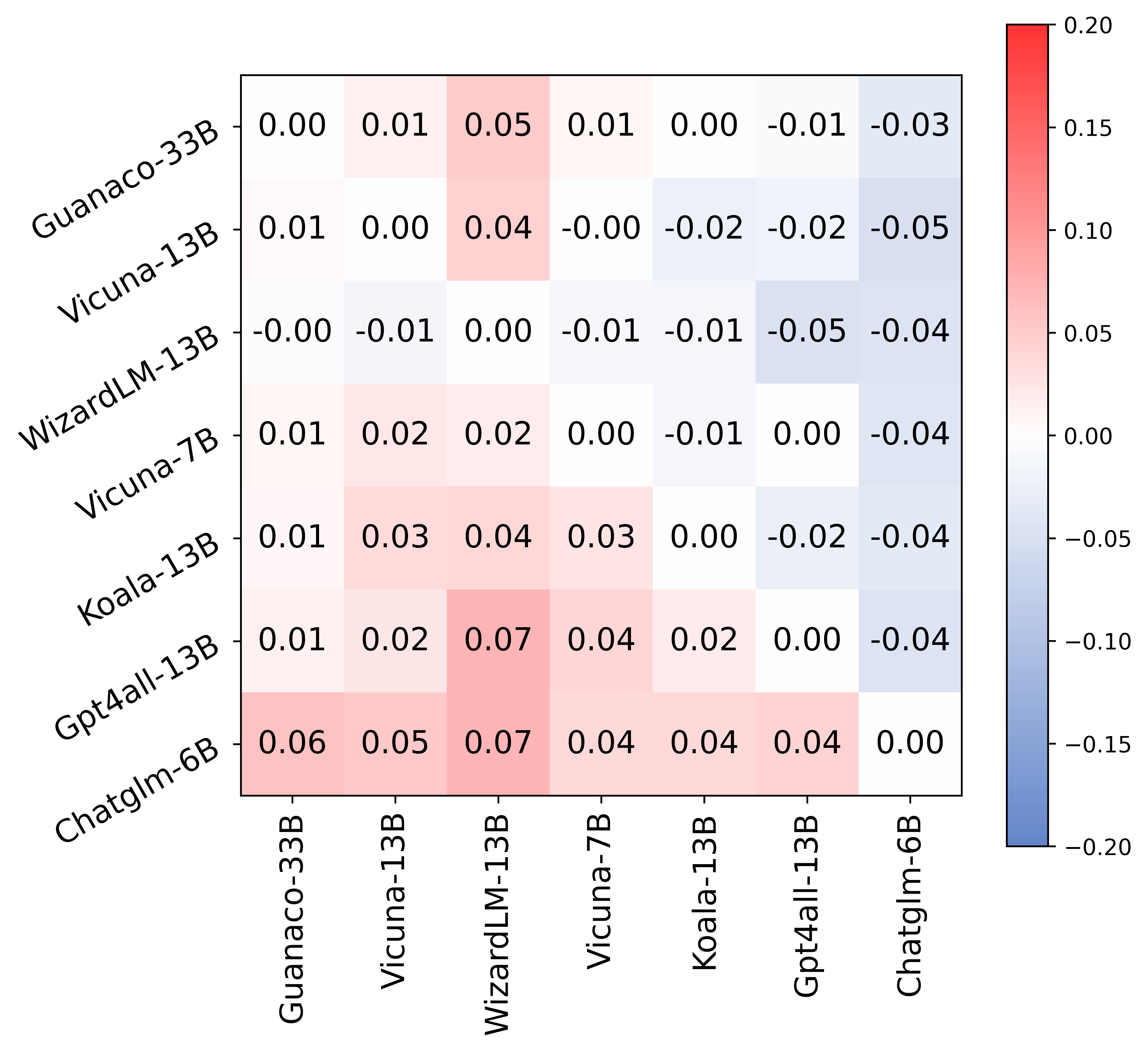}
		\caption{AlpacaEval (weighted PG)}
		\label{chutian6} 
	\end{subfigure}
	\caption{Heatmap distribution of preference gap (PG) metric among seven LLMs across three datasets. Higher values (above 0) indicate greater evaluation bias. The first row shows original PG values in three datasets, while the second row displays PG values re-weighted using our learned confidence weights.}
	\label{fig:hotmap}
\end{figure*}

\subsection{Exploring the Role of Confidence Weight}


In this subsection, we show that the confidence weight $w$ learned by our \textit{consistency optimization} can reduce the system evaluation bias. Specifically, we first study whether the ``review'' model would prefer a particular model's response. Following \citep{chu2024pre}, we employ the preference gap (PG) to evaluate the bias as follows, 
\vspace*{-0.5\baselineskip} 
\begin{equation}
\label{eq:PG}
PG(i,j)=P_i(i>j)-P_j(i>j),
\end{equation}
where $P_i(i>j)$ represents the winning rate of model $i$ as the ``reviewer'' believes that $i$ defeated $j$. 
The heatmap distribution of the PG value $PG(i,j)$ among seven LLMs across three datasets is demonstrated in the first row of Figure \ref{fig:hotmap}. It can be observed that the evaluation system exhibits severe bias. Especially on ChatGLM-6B and Mpt-7B models, they often believe that their results are better than other ones, as their PG values are greater than 0 across three datasets.

After the \textit{consistency optimization}, we assign the learned confidence weight $w$ to the corresponding model and ultimately obtain the re-weighting PG value $\hat{PG}(i, j)$ as follows, 
\vspace*{-0.5\baselineskip} 
\begin{equation}
\label{eq:PG'}
\hat{PG}(i,j)=w_i\times P_i(i>j)-w_j\times P_j(i>j).
\end{equation}
The results of the re-weighting PG value $\hat{PG}(i, j)$ are displayed on the second row of Figure \ref{fig:hotmap}. It can be observed that the learned confidence weight $w$ can significantly mitigate the preference gaps of the whole evaluation system. In our consistency optimization, LLMs such as ChatGLM-6B and Mpt-7B have lower weights, and reducing their confidence can effectively alleviate the system evaluation bias.

\begin{figure*}[!t]
    \centering
    \includegraphics[width=1\textwidth]{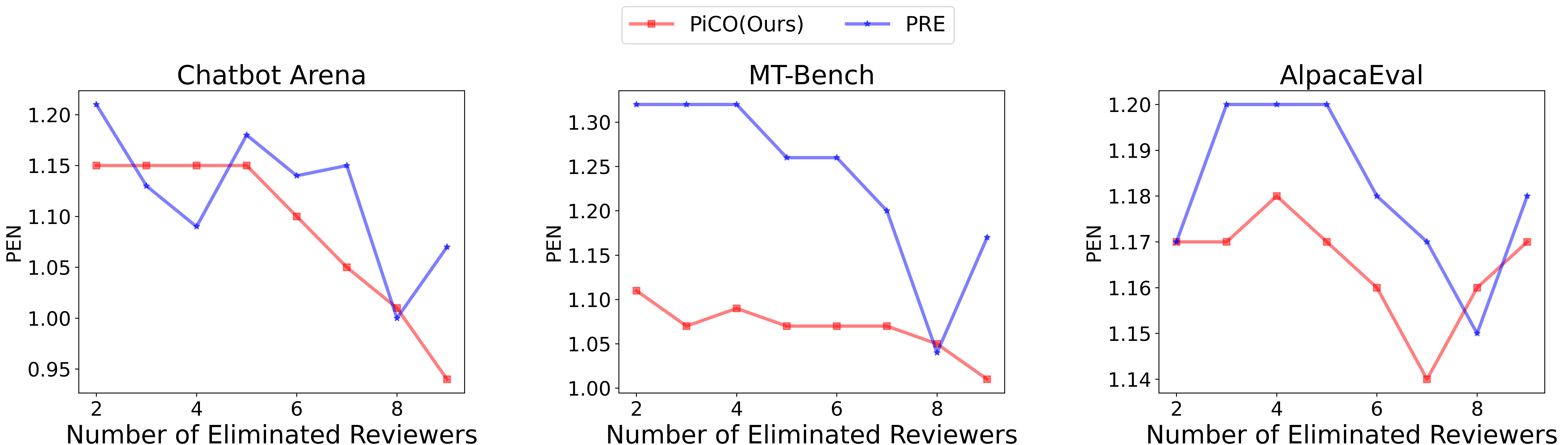}
    \caption{Performance comparison of the PiCO (Ours) and PRE methods on the Chatbot Arena, MT-Bench, and AlpacaEval datasets, with the number of eliminated reviewers on the x-axis. The y-axis is PEN, where lower values indicate better performance.}
    \label{fig:plot}
    \vspace{-0.35cm}
\end{figure*}
\subsection{Study of Elimination Mechanism}
\label{section:3.4}


\begin{wrapfigure}{r}{0.35\textwidth}
    \centering
    \vspace{-0.3cm}
    \includegraphics[width=1.0\linewidth]{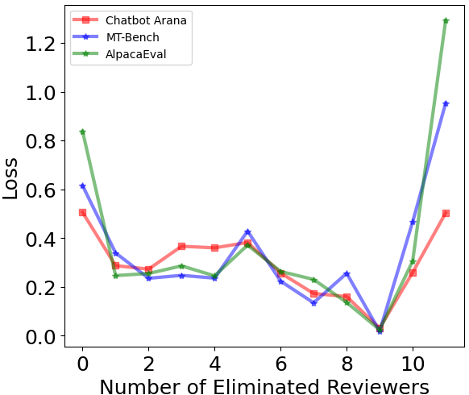}
    \caption{The average loss for different numbers of eliminated reviewers($\downarrow$). It shows how the iterative elimination of weaker reviewers affects the overall loss in the peer-review system.}
    \vspace{-0.3cm}
    \label{fig:loss}
\end{wrapfigure}

\textbf{Performance Comparison of Elimination Mechanisms.} The PiCO and PRE methods both employ elimination mechanisms to remove those weakest LLMs from the ``reviewer queue'' during the evaluation process. As shown in Figure \ref{fig:plot}, the x-axis quantifies the number of reviewers eliminated, and the y-axis measures the PEN, where lower scores denote higher performance. It can be observed that both PiCO and PRE exhibit better performance with an increasing number of eliminated ``reviewers''. The proposed PiCO approach can achieve better performance than PRE in most cases. It is worth noting that the PRE method employs the accuracy of ``qualification exams'' to eliminate weak LLMs, and this process requires human annotation \citep{chu2024pre}. On the contrary, the elimination process of our PiCO method is unsupervised and can still achieve better evaluation results than PRE.

\textbf{Automatic Learning of Elimination Thresholds.} We observed that weaker LLMs tend to have poorer evaluation abilities, introducing significant noise into the peer-review system. Therefore, eliminating weaker models instead of retaining them enhances the robustness of the system. We employed an unsupervised approach to automatically learn the elimination threshold, as shown in Figure \ref{fig:loss}, by using the average training loss curve as the number of eliminated reviewers increases. It can be seen that removing weaker reviewers reduces the average loss of the entire system, indicating that eliminating noisy evaluations benefits the overall process. Notably, when 60$\%$ (or 9) of the weaker reviewers are removed, the system's loss reaches its minimum. This trend is consistent across all three datasets, suggesting that the elimination threshold is learned automatically. However, removing more than 9 stronger reviewers harms the evaluation process.

\subsection{Other Results}


\textbf{Validation on more metrics (Precision@K and RBP@K).} We demonstrated the results of precision and RBP (K=8,9,10) with other baselines in Table \ref{table:3} (left). The results show that the proposed PiCO approach can achieve better precision and RBP performance in all cases. These results once again validate that PiCO can predict the LLM ranking more accurately than other baselines. 

\textbf{Comparison of tokens consumed.} We compute the token consumption of each method in Table \ref{table:3} (right). It can be observed that the proposed PiCO approach has a similar token consumed with other baselines (\textit{e.g.}, PRD and PRE) while achieving better evaluation performance. Although Chatbot Arena has a smaller token consumption, it requires 33k human annotations, while PiCO does not require any human annotations.

\begin{figure*}[!t]
    \centering
    \includegraphics[width=0.99\textwidth]{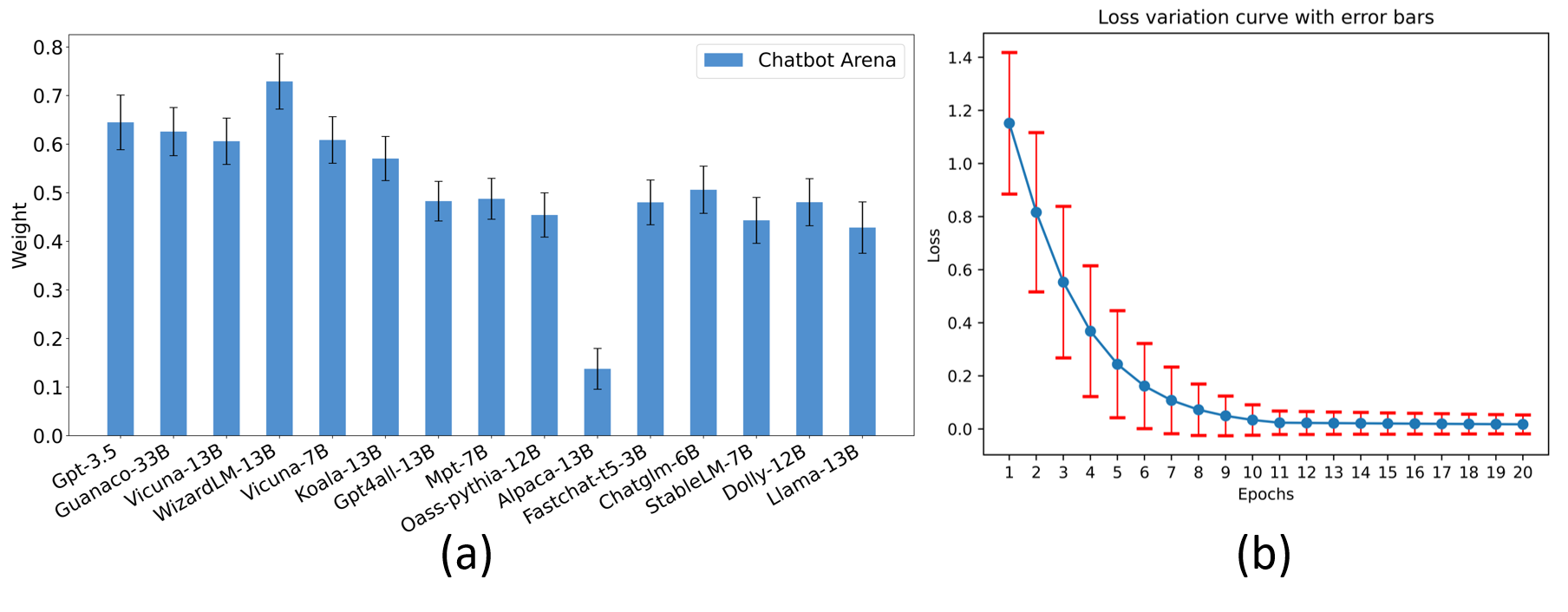}
    \caption{Stability validation of consistency optimization. We repeated the experiment with different seeds for 1000 times, and plotted the training loss curve and weight distribution. The results show that the learning process is stable and the learned $w$ is convergence. 
    }
    \label{fig:5}
\end{figure*}

\textbf{Stability validation of consistency optimization.} We repeated the experiment with different seeds for 1000 times, and plotted the training loss curve and weight distribution in Figure \ref{fig:5}. The results show that the proposed consistency optimization process is stable and the learned $w$ is convergence. 

\textbf{Comparing with existing benchmarks.} We select the widely-used benchmarks (i.e., MMLU \citep{hendrycks2020measuring} and GSM8K \citep{cobbe2021training}) to evaluate the model performance ranking $\hat{\mathcal{R}}$, and calculate the Spearman's $S(\uparrow)$ and Kendall's $\tau(\uparrow)$ rank correlation with the human preference ranking $\mathcal{R}^*$. The results are demonstrated in Table \ref{table:4}. It can be observed that these benchmarks can only measure LLMs’ specific capability on a confined set of tasks, which fails to assess their alignment with human preference. These phenomena have been widely validated in other literature \citep{zhou2023don,zheng2023judging,chang2023survey} and have almost become a consensus in the community of LLM evaluation.


\begin{table*}[!t]
    \renewcommand{\arraystretch}{1}
    \setlength{\tabcolsep}{1pt}
    \centering
    \caption{Comparison of more metrics (Precision@K and RBP@K) and token consumption on Chatbot Arena.}
    \label{table:3}
    \resizebox{\textwidth}{!}{
    \begin{tabular}{c | c c c | c c c | c c c}
        \toprule
        \multirow{2}{*}{Methods} & \multicolumn{3}{c}{RBP@K $(\uparrow)$} & \multicolumn{3}{|c|}{Precision@K $(\uparrow)$} & \multirow{2}{*}{Input Token} & \multirow{2}{*}{Output Token} & \multirow{2}{*}{Annotation Cost} \\ 
        ~ & 8 & 9 & 10 & 8 & 9 & 10 & ~\\
        \hline
        Chatbot Arena Platforms & - & - & - & - & - & - & $\sim7500$k & $\sim10944$k & $\sim32$k \\
        GPTScore(flan-t5-xxl) & 26.2\% & 29.6\% & 45.1\% & 50.0\% & 55.6\% & 70.0\% & $\sim22882$k & $\sim12260$k & 0 \\
        GPTScore(davinci-002) & 42.0\% & 50.6\% & 53.3\% & 62.5\% & 77.8\% & 80.0\% & $\sim22882$k & $\sim12260$k & 0 \\
        PandaLM & 63.5\% & 63.5\% & 66.2\% & 62.5\% & 55.6\% & 60.0\% & $\sim22882$k & $\sim10355$k & 0 \\
        PRD & 67.2\% & 73.8\% & 81.3\% & 87.5\% & 88.9\% & 80.0\% & $\sim25087$k & $\sim10935$k & 0 \\
        PRE & 78.0\% & 81.3\% & 81.3\% & 87.5\% & 88.9\% & 80.0\% & $\sim24120$k & $\sim11115$k & $\sim7$k \\
        \textbf{PiCO (Ours)} & \textbf{83.2\%} & \textbf{83.2\%} & \textbf{85.9\%} & \textbf{100.0\%} & \textbf{100.0\%} & \textbf{90.0\%} & $\sim23823$k & $\sim11685$k & 0 \\
        
        \toprule
    \end{tabular}
    }
    \vspace{-0.35cm}
\end{table*}

\section{Related Work}

\textbf{Evaluation Benchmarks for Diversity.}
LLMs are designed to handle a variety of tasks, necessitating comprehensive benchmarks \citep{chang2023survey}. Notable benchmarks include GLUE and SuperGLUE \citep{wang2018glue,wang2019superglue}, which simulate real-world scenarios across tasks such as text classification, translation, reading comprehension, and dialogue generation. HELM \citep{liang2022holistic} provides a holistic evaluation of LLMs, assessing language understanding, generation, coherence, and reasoning. BIG-bench \citep{srivastava2022beyond} pushes LLM capabilities with 204 diverse tasks. MMLU \citep{hendrycks2020measuring} measures multitask accuracy across domains like mathematics and law. GSM8K \citep{cobbe2021training} including 8K simple math questions with detailed solutions is widely used to assess the mathematical reasoning of models on grade-school-level questions. However, these evaluations can be compromised by benchmark leakage, where evaluation data inadvertently used for training leads to inflated performance metrics \citep{aiyappa2023trust}.

\begin{table*}[!t]
    \renewcommand{\arraystretch}{1}
    \setlength{\tabcolsep}{1pt}
    \centering
    \caption{Comparison with existing benchmarks.}
    \label{table:4}
    \begin{tabular}{c | c c }
        \toprule
        \multirow{2}{*}{Benchmarks} & Spearman's Rank & Kendall's Rank \\ 
        & Correlation Coefficient $S(\uparrow)$ & Correlation Coefficient $\tau(\uparrow)$ \\
        \hline
        MMLU & 0.53 & 0.37 \\
        GSM8K & 0.32 & 0.15 \\
        \textbf{PiCO (Ours)} & \textbf{0.88} & \textbf{0.67} \\
        \toprule
    \end{tabular}
\end{table*}

\textbf{Human Evaluation.}
Human evaluation provides reliable feedback that closely aligns with real-world applications \citep{chang2023survey}. \cite{liang2022holistic} evaluated summary and misinformation scenarios across multiple models. \cite{ziems2023can} involved experts to assess model outputs in various domain-specific tasks. \cite{bang2023multitask} examined ChatGPT's performance in summarization, translation, and reasoning using human-annotated datasets. The LMSYS initiative introduced platforms like Chatbot Arena \citep{zheng2023judging}, relying on human ratings as the primary evaluation metric. Currently, using these anonymous battle platforms has become the primary way to evaluate the large language models, and its success is attributed to the wisdom of the crowds \citep{surowiecki2005wisdom,budescu2015identifying,weller2007cultural} and have been proven to lead to better results than that of an individual. 
Despite its effectiveness, human evaluation is costly and subject to bias and cultural differences\citep{peng1997validity}.

\textbf{Large Language Models for Evaluation.}
The trend towards developing open-source LLMs has led to initiatives employing one or multiple LLMs as evaluators for assessing the outputs of LLMs. GPTScore \citep{fu2023gptscore} uses models like GPT-3 to assign probabilities to high-quality content through multidimensional evaluation. \cite{bubeck2023sparks} tested GPT-4, finding it rivaling human capabilities. Lin and Chen introduced LLM-EVAL \citep{lin2023llm} for evaluating dialogue quality with single prompts. PandaLM \citep{wang2023pandalm} employs LLMs as "judges" for evaluating instruction tuning. However, reliance on a single model can introduce biases such as positional \citep{dettmers2024qlora}, verbosity \citep{wang2024far}, and self-favoring biases \citep{liu2023gpteval,zheng2023judging}. ChatEval \citep{chan2023chateval} proposes a multi-agent framework to simulate human evaluation processes. Similarly, PRE \citep{chu2024pre} and PRD \citep{li2023prd} use LLMs as evaluators, combining multiple evaluation outcomes for automated assessment. However, like the PRE method, which uses human feedback for supervised evaluation throughout the process, the comprehensive assessment of LLMs still incurs a relatively high cost.

\section{Conclusion}
In this paper, we propose PiCO, a novel unsupervised evaluation method to automatically evaluate Large Language Models (LLMs) without relying on human feedback. PiCO utilizes \textit{peer-review} mechanisms to autonomously assess LLMs in a shared environment, where both open-source and closed-source models can respond to unlabeled questions and evaluate each other. In this setup, each LLM's response score is determined collectively by other anonymous models, aiming to maximize consistency across capabilities and scores. The extensive experiment results across multiple datasets and standard rank-based metrics demonstrate that PiCO effectively generates an LLM ranking that aligns closely with human preferences. In the future, we plan to extend the peer-review mechanism to evaluate the capabilities of multi-modality large models.

\section*{Acknowledgements}
This work was supported in part by the Natural Science Foundation of China (No. 62202014, 62332002, 62425101, 62088102), and also supported by the China Postdoctoral Science Foundation under Grant Number BX20240013 and 2024M760113.

\bibliography{iclr2025_conference}
\bibliographystyle{iclr2025_conference}

\clearpage
\appendix

\section{Detailed Explanation of Metrics}
\label{appendix:Metric}
In this section, we provide a comprehensive explanation of the metrics used to evaluate the alignment between learned LLM rankings and human rankings. These metrics assess the strength of correlations, complexity, and the level of agreement between rankings. Specifically, we discuss five key metrics: Spearman's Rank Correlation Coefficient, Kendall's Rank Correlation Coefficient, Permutation Entropy, Count Inversions, and Longest Increasing Subsequence, detailing their formulations and intuitive interpretations.

\textit{i)} \textbf{Spearman's Rank Correlation Coefficient $S(\uparrow)$} \citep{lehman2013jmp} measures the strength and direction of the monotonic relationship between two ranked variables. It is computed as:
\begin{equation}
    S(\hat{\mathcal{R}}, \mathcal{R}^*) = 1 - \frac{6 \sum_{i=1}^{m} d_i^2}{m(m^2-1)},
\end{equation}
where $d_i = \text{rank}_{\hat{\mathcal{R}}}(M_i) - \text{rank}_{\mathcal{R}^*}(M_i)$ is the difference between the ranks of LLM $M_i$ in the learned ranking $\hat{\mathcal{R}}$ and the human ranking $\mathcal{R}^*$, and $m$ is the total number of LLMs. A higher Spearman coefficient indicates a stronger correlation between the rankings.

\textit{ii)} \textbf{Kendall's Rank Correlation Coefficient $\tau(\uparrow)$} \citep{kendall1938new} evaluates the similarity between two rankings by counting the number of concordant and discordant pairs. It is given by:
\begin{equation}
    \tau(\hat{\mathcal{R}}, \mathcal{R}^*) = \frac{C - D}{\frac{1}{2}m(m-1)},
\end{equation}
where $C$ represents the number of concordant pairs, and $D$ represents the number of discordant pairs. A pair $(M_i, M_j)$ is concordant if $M_i$ and $M_j$ have the same order in both $\hat{\mathcal{R}}$ and $\mathcal{R}^*$, meaning if $M_i \succ M_j$ in $\hat{\mathcal{R}}$, then $M_i \succ M_j$ in $\mathcal{R}^*$. Conversely, a pair is discordant if their relative order differs between the two rankings. A higher $\tau$ value indicates a closer alignment between the rankings.

\textit{iii)} \textbf{Permutation Entropy $H(\downarrow)$} \citep{bandt2002permutation} measures the complexity or randomness of sequences, which is formulated as follows:
\begin{equation}
\label{eq:PEN}
    H(\hat{\mathcal{R}}, \mathcal{R}^*) := -\sum p(\pi)\log p(\pi),
\end{equation}
where
\begin{equation*}
    p(\pi)=\frac{\#\{t|0\leq t \leq m-k, (M_{t+1},...,M_{t+k}) \in \pi\}}{m-k+1}.
\end{equation*}
$\pi$ denotes different permutations, $k$ is a hyper-parameter recommended to be set to 3 to 7, and we set $k=3$ in this paper. Intuitively, it samples some subsequences and calculates the entropy for all permutation types. And the lower the permutation entropy in the learned LLM rankings, the closer it is to the ground-truth human rankings.

\textit{iv)} \textbf{Count Inversions $C(\downarrow)$}. Counting inversions \citep{leiserson1994introduction} aims to measure the degree of disorder or "invertedness" in an array or sequence of elements. We thus define it as follows, 
\begin{equation}
\label{eq:CIN}
    C(\hat{\mathcal{R}}, \mathcal{R}^*) := \sum_{M_i, M_j \sim \mathcal{M}}\mathbf{1}\{M_i \succ M_j \land i < j\}.
\end{equation}
Where $\mathbf{1}\{\cdot\}$ is the indicator function that the value is 1 when the condition is met, otherwise it is 0. Intuitively, the fewer inverse pairs in the learned LLM rankings, the closer it is to the ground-truth human rankings. 

\textit{v)} \textbf{Longest Increasing Subsequence $L(\uparrow)$}. The longest increasing subsequence aims to find the length of the longest subsequence in a given sequence of elements, where the subsequence is in increasing order. We utilize it to measure the degree of match with human rankings as follows, 
\begin{equation}
\label{eq:LIS}
    L(\hat{\mathcal{R}},\mathcal{R}^*):=\max\left\{dp[i]\mid1\leq i\leq m\right\},
\end{equation}
where
\begin{equation*}
    dp[i]=1+\max\left\{dp[j]\mid1\leq j<i \land M_j \prec M_i  \right\}.
\end{equation*}
$dp[i]$ represents the length of the longest increasing subsequence that ends with $M_i$. LIS allows for a nuanced understanding of the degree to which the learned ranking aligns with the ideal human ranking, with a higher LIS length indicating greater alignment.

\section{Dataset Format}

\begin{figure*}[!ht]
    \centering
    \includegraphics[width=0.8\textwidth]{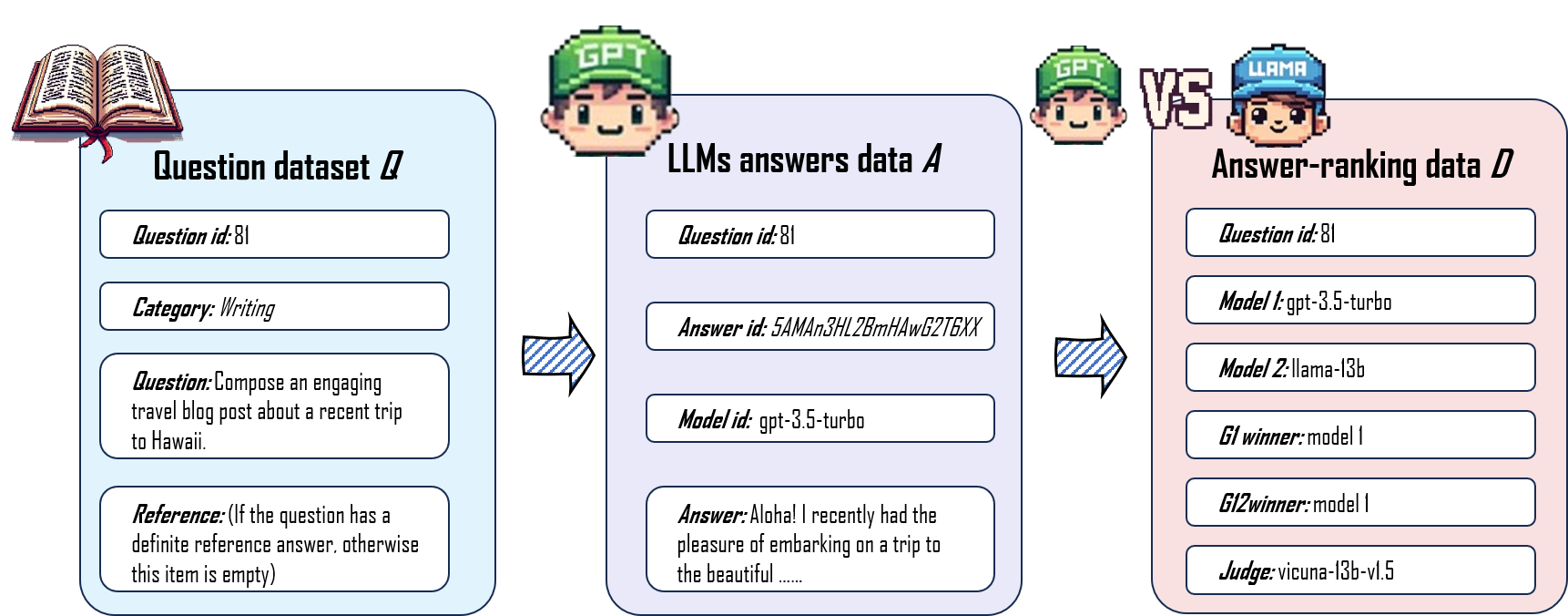}
    \caption{Format of the Question dataset $\mathcal{Q}$, LLMs responses data $\mathcal{A}$, and the Answer-Ranking data $\mathcal{D}$ for Peer Review}
    \label{fig:dataset}
\end{figure*}

\ys{Focusing on the MT-Bench dataset, we demonstrate the ensuing data format utilizing dataset $\mathcal{Q}$. As Figure \ref{fig:dataset} illustrates, the Question dataset $\mathcal{Q}$ contains "Question id," "Category," "Question," and "Reference." In categories with definitive answers like "reasoning" or "math," the "Reference" field is populated with standard answers; otherwise, it remains blank. Each model M in our pool processes the Question dataset $\mathcal{Q}$ to generate the LLMs answer data $\mathcal{A}$, consisting of "Question id," "Answer id," "Model id," and "Answer." Finally, we combine pairs in $\mathcal{A}$ and appoint judges to evaluate, creating the Answer-Ranking data $\mathcal{D}$, featuring "Question id," "Model 1," "Model 2," "G1 winner," "G2 winner," and "Judge." Here, "G1 winner" and "G2 winner" indicate the outcomes of inputting reversed order responses of Model 1 and Model 2 into the judge model, a method employed to mitigate biases stemming from models' preferences for input order.}

\section{Detailed Prompt for Reviewers}
\label{appendix:B}

The evaluation prompts, as detailed in Section 2.2.1, are employed during the Peer Review Stage. These prompts are provided to the Reviewer Language Model Systems (LLMs), enabling them to generate evaluative preferences. In our experimental framework, we devised four distinct prompt settings. For each setting, a tailored prompt template was meticulously crafted as illustrated below:

\textbf{Template for Single-Turn Interaction: }This template is designed for single-turn interactions between users and LLMs, where there is no predetermined correct answer. It facilitates open-ended dialogue, allowing for a wide range of user inquiries without the expectation of specific responses.

\textbf{Referenced Template for Single-Turn Interaction:} Tailored for single-turn dialogues between users and LLMs, this template incorporates predefined correct answers. It is particularly suited for interactions involving factual inquiries, such as mathematics or logic problems, where accuracy and reference to correct information are paramount.

\textbf{Template for Multi-Turn Interaction: }This template caters to multi-turn conversations between users and LLMs, without predefined answers. It supports extended interactions, enabling users to explore topics in depth through a series of interconnected questions and responses.

\textbf{Referenced Template for Multi-Turn Interaction:} Designed for multi-turn dialogues with predefined correct answers, this template is ideal for complex inquiries requiring sequential reasoning or problem-solving, such as mathematical computations or logical deductions. 

Each template is carefully constructed to match its intended use-case, providing a structured framework that guides the interaction between users and LLMs towards achieving desired outcomes, whether for open-ended exploration or precise problem-solving.

\begin{tcolorbox}[colback=gray!20!white, colbacktitle=white, coltitle=black, colframe=black!75!black, boxrule=0.7pt, halign title=center, title=\textbf{Template for Single-Turn Answer}]
\textbf{System prompt:} Please act as a judge and evaluate the quality of the responses provided by two AI assistants to the user question displayed below. You do not need to explain, just give your judgment. Output your final verdict by strictly following this format: "[[A]]" if assistant A is better, "[[B]]" if assistant B is better, and "[[C]]" for a tie.

\textbf{User Question:} \{question\}

\textbf{Assistant A's Answer: } \{answer a\}

\textbf{Assistant B's Answer: } \{answer b\}

\end{tcolorbox}
\begin{tcolorbox}[colback=gray!20!white, colbacktitle=white, coltitle=black, colframe=black!75!black, boxrule=0.7pt, halign title=center, title=\textbf{Referenced Template for Single-Turn Answer}]
\textbf{System prompt:} Please act as a judge and evaluate the quality of the responses provided by two AI assistants to the user question displayed below, with reference to the provided reference answers. You do not need to explain, just give your judgment. Output your final verdict by strictly following this format: "[[A]]"if assistant A is better, "[[B]]" if assistant B is better, and "[[C]]" for a tie.

\textbf{User Question:} \{question\}

\textbf{Reference Answer: } \{reference answer\}

\textbf{Assistant A's Answer: } \{answer a\}

\textbf{Assistant B's Answer: } \{answer b\}

\end{tcolorbox}

\begin{tcolorbox}[colback=gray!20!white, colbacktitle=white, coltitle=black, colframe=black!75!black, boxrule=0.7pt, halign title=center, title=\textbf{Template for Multi-Turn Answer}]
\textbf{System prompt:} Please act as a judge and evaluate the quality of the responses provided by two AI assistants to the user question displayed below. You do not need to explain, just give your judgment. Output your final verdict by strictly following this format: "[[A]]" if assistant A is better, "[[B]]" if assistant B is better, and "[[C]]" for a tie

\textbf{Assistant A's Conversation with User:} 

\qquad \textbf{User: } \{question 1\}

\qquad \textbf{Assistant A: } \{answer a1\}

\qquad \textbf{User: } \{question 2\}

\qquad \textbf{Assistant A: } \{answer a2\}

\textbf{Assistant B's Conversation with User:} 

\qquad \textbf{User: } \{question 1\}

\qquad \textbf{Assistant B: } \{answer b1\}

\qquad \textbf{User: } \{question 2\}

\qquad \textbf{Assistant B: } \{answer b2\}

\end{tcolorbox}

\begin{tcolorbox}[colback=gray!20!white, colbacktitle=white, coltitle=black, colframe=black!75!black, boxrule=0.7pt, halign title=center, title=\textbf{Referenced Template for Multi-Turn Answer}]
\textbf{System prompt:} Please act as a judge and evaluate the quality of the responses provided by two AI assistants to the user question displayed below, in comparison to the reference answers. You do not need to explain, just give your judgment. Output your final verdict by strictly following this format: "[[A]]"if assistant A is better, "[[B]]" if assistant B is better, and "[[C]]" for a tie.

\textbf{Reference Answer} 

\qquad \textbf{User: } \{question 1\}

\qquad \textbf{Reference answer: } \{ref answer 1\}

\qquad \textbf{User: } \{question 2\}

\qquad \textbf{Reference answer: } \{ref answer 2\}

\textbf{Assistant A's Conversation with User:} 

\qquad \textbf{User: } \{question 1\}

\qquad \textbf{Assistant A: } \{answer a1\}

\qquad \textbf{User: } \{question 2\}

\qquad \textbf{Assistant A: } \{answer a2\}

\textbf{Assistant B's Conversation with User:} 

\qquad \textbf{User: } \{question 1\}

\qquad \textbf{Assistant B: } \{answer b1\}

\qquad \textbf{User: } \{question 2\}

\qquad \textbf{Assistant B: } \{answer b2\}
\end{tcolorbox}

\section{Scoring Methodology}
\label{appendix:C}
In Section 2.2.2, Equation \ref{eq:G} delineates the methodology for optimizing scores. Within this framework, the function $\mathbf{1}{\{A_i^j > A_i^k\}}$ is more precisely defined as $f(A_i^j, A_i^k)$. Additionally, the function $f(A_i^j, A_i^k)$ is not fixed and can be implemented using various computational strategies. We introduce two distinct methodologies in this context: the Elo mechanism and the Rank mechanism.

Within the framework of the Elo mechanism, as specified by Equation \ref{eq: cal_score_1}, the $BASE$ value is set to 10, and the $SCALE$ factor is determined to be 400. This approach facilitates a dynamic adjustment of scores based on the outcomes of pairwise comparisons, allowing for a nuanced reflection of performance variations among models.

Conversely, in the context of the Rank mechanism, as outlined by Equation \ref{eq: cal_score_2}, $rank(j)$ signifies the current ranking of model $j$, with the constant $K$ assigned a value of 200. This mechanism employs a model's ranking within a predefined hierarchy as a pivotal factor in score calculation, thereby providing a straightforward, yet effective, method for evaluating comparative model performance.

\begin{equation}
\label{eq: cal_score_1}
f(A_i^j,A_i^k)=
\begin{cases}
1- \frac{1}{1 + \text{BASE}^{((G(k) - G(j)) / \text{SCALE})}} &\text{if }A_i^j>A_i^k\\
0.5- \frac{1}{1 + \text{BASE}^{((G(k) - G(j)) / \text{SCALE})}}&\text{if }A_i^j=A_i^k\\
0- \frac{1}{1 + \text{BASE}^{((G(k) - G(j)) / \text{SCALE})}}&\text{if }A_i^j<A_i^k
\end{cases}
\end{equation}

\begin{equation}
\label{eq: cal_score_2}
f(A_i^j,A_i^k)=\begin{cases}1+(rank(j)-rank(k))/K&\text{if }A_i^j>A_i^k\\
0.5&\text{if }A_i^j=A_i^k\\
0&\text{if }A_i^j<A_i^k
\end{cases}
\end{equation}

\section{Overall Algorithm of Peer Review}
\label{appendix:D}

The overall algorithm, as delineated in Algorithm \ref{algorithm 2}, encapsulates the comprehensive process outlined in Section 2.2. This sequence commences with "Data Collection and LLMs Pool Construction," progresses through "Answer-Ranking Data Construction Based on Peer Review," advances to "Consistency Optimization," and culminates with the "Unsupervised Elimination Mechanism."

\begin{algorithm}[t]
\caption{Overall Framework Algorithm of Peer Review}
\label{algorithm 2}
\begin{algorithmic}[1]
\Require Unlabeled dataset $\mathcal{Q}$, Pool of LLMs $\mathcal{M}$, Active LLM pool $\mathcal{M}^* = \mathcal{M}$
\Ensure Consistency-optimized ranking of LLMs $\mathcal{R}^*$

\State Initialize response matrix $A \gets \emptyset$
\For{each question $q_i \in \mathcal{Q}$}
    \State Initialize response vector for question $q_i$, $A^i \gets \emptyset$
    \For{each model $m_j \in \mathcal{M}$}
        \State $A^i_j \gets \text{response of model } m_j \text{ to question } q_i$
        \State $A^i \gets A^i \cup \{A^i_j\}$
    \EndFor
    \State Shuffle $A^i$ to obtain permuted response vector $A^i$
    \State $A \gets A \cup \{A^i\}$
\EndFor

\State Initialize answer-ranking data $D \gets \emptyset$
\State Initialize model weights vector $w$ with Gaussian distribution
\For{each permuted response vector $A^i$}
    \For{each pair of responses $(A^j_i, A^k_i)$ in $A^i$}
        \For{$s \gets 1$ to $5$} \Comment{Randomly select 5 models for evaluation}
            \State Evaluate the pair $(A^j_i, A^k_i)$ with model $m_s$
            \State $D \gets D \cup \{(A^j_i, A^k_i, > w^s)\}$
        \EndFor
    \EndFor
\EndFor

\State Initialize scores $G_j$ for each model $m_j \in \mathcal{M}$ to the Elo initial score

\Repeat
    \While{not converged}
        \For{each model $m_j \in \mathcal{M}$}
            \State Compute $G_j$ using updated formula:
            \State $G_j = \sum_{i}\sum_{k \neq j}\sum_{s \neq k, s \neq j} \mathbf{1} \{A_i^j, A_i^k\} \times w^s \quad (A^j_i, A^k_i, > w^s, s \in \mathcal{M}^*) \in D$
        \EndFor
        \State Update weight vector $w$ to maximize the consistency of $w$ and $G$
    \EndWhile
    \State Sort $\mathcal{M}^*$ by $G_j$ to identify $\mathcal{M}_{min}$, the lowest-scoring model
    \If{size of $\mathcal{M}^*$ > threshold}
        \State Remove $\mathcal{M}_{min}$ from $\mathcal{M}^*$
    \EndIf
\Until{size of $\mathcal{M}^*$ < threshold}

\State Compute the final ranking $\mathcal{R}^*$ based on the optimized scores $G_j$
\State \textbf{return} $\mathcal{R}^*$
\end{algorithmic}
\end{algorithm}

\section{Complete Experimental Results}
\label{appendix:E}

\begin{figure*}[!ht]
    \centering
    \includegraphics[width=1\textwidth]{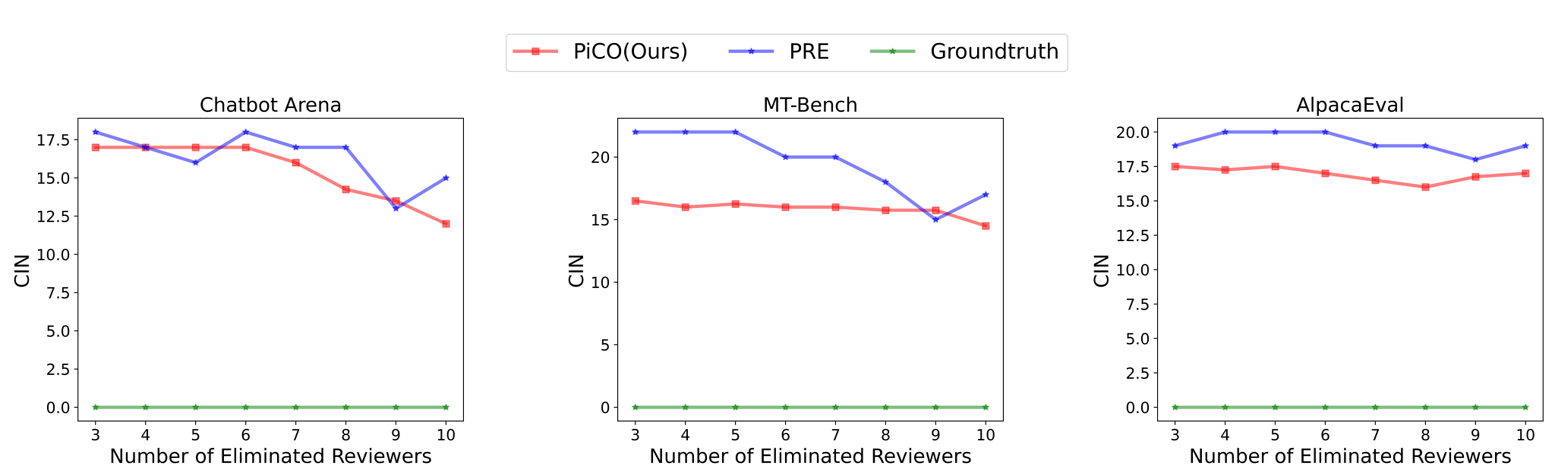}
    \caption{Performance comparison of the PiCO (Ours) and PRE \citep{chu2024pre} methods on the MT-Bench, Chatbot Arena, and AlpacaEval datasets, with the number of eliminated reviewers on the x-axis. The y-axis is CIN, where lower values indicate better performance.}
    \label{fig:PEN_plot}
\end{figure*}

\begin{figure*}[!ht]
    \centering
    \includegraphics[width=1\textwidth]{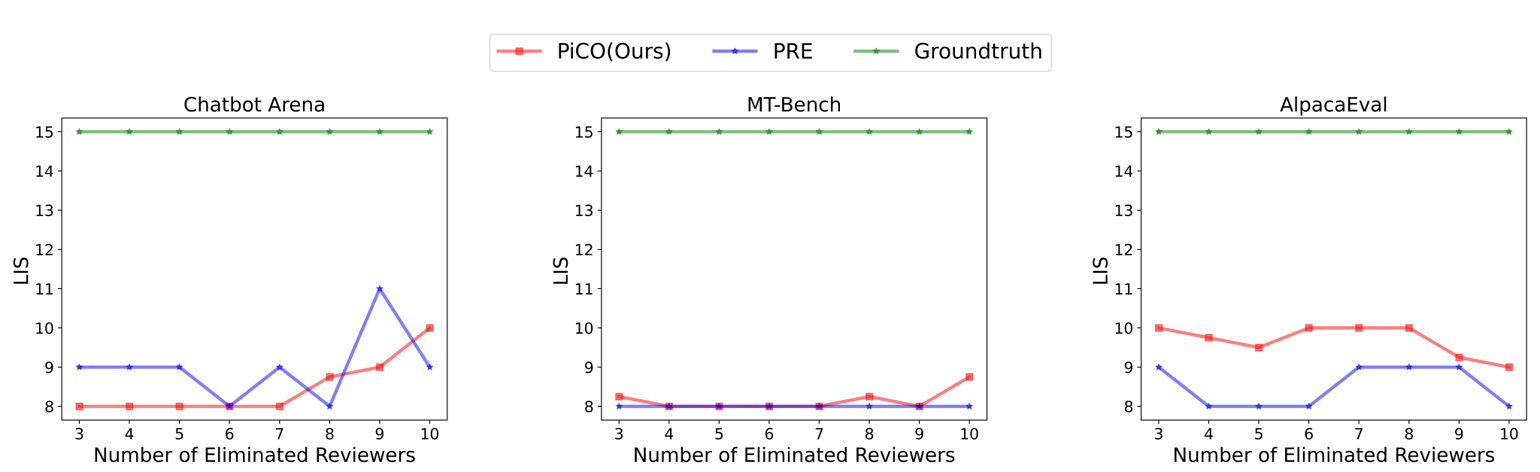}
    \caption{Performance comparison of the PiCO (Ours) and PRE \citep{chu2024pre} methods on the MT-Bench, Chatbot Arena, and AlpacaEval datasets, with the number of eliminated reviewers on the x-axis. The y-axis is LIS, where upper values indicate better performance.}
    \label{fig:LIS_plot}
\end{figure*}

In Section 3.4, we both employ elimination mechanisms to cull the weakest LLMs from the 'reviewer queue' during the evaluation process. In Figures \ref{fig:PEN_plot} and \ref{fig:LIS_plot}, we present the results for the PEN and LIS metrics, where lower PEN scores indicate better performance, and higher LIS scores denote superior performance. It is evident that both the 'PiCO' and PRE approaches demonstrate enhanced performance as the number of eliminated 'reviewers' increases. In most cases, the proposed 'PiCO' method outperforms PRE.

\begin{figure*}[ht]
    \centering
    \includegraphics[width=1\textwidth]{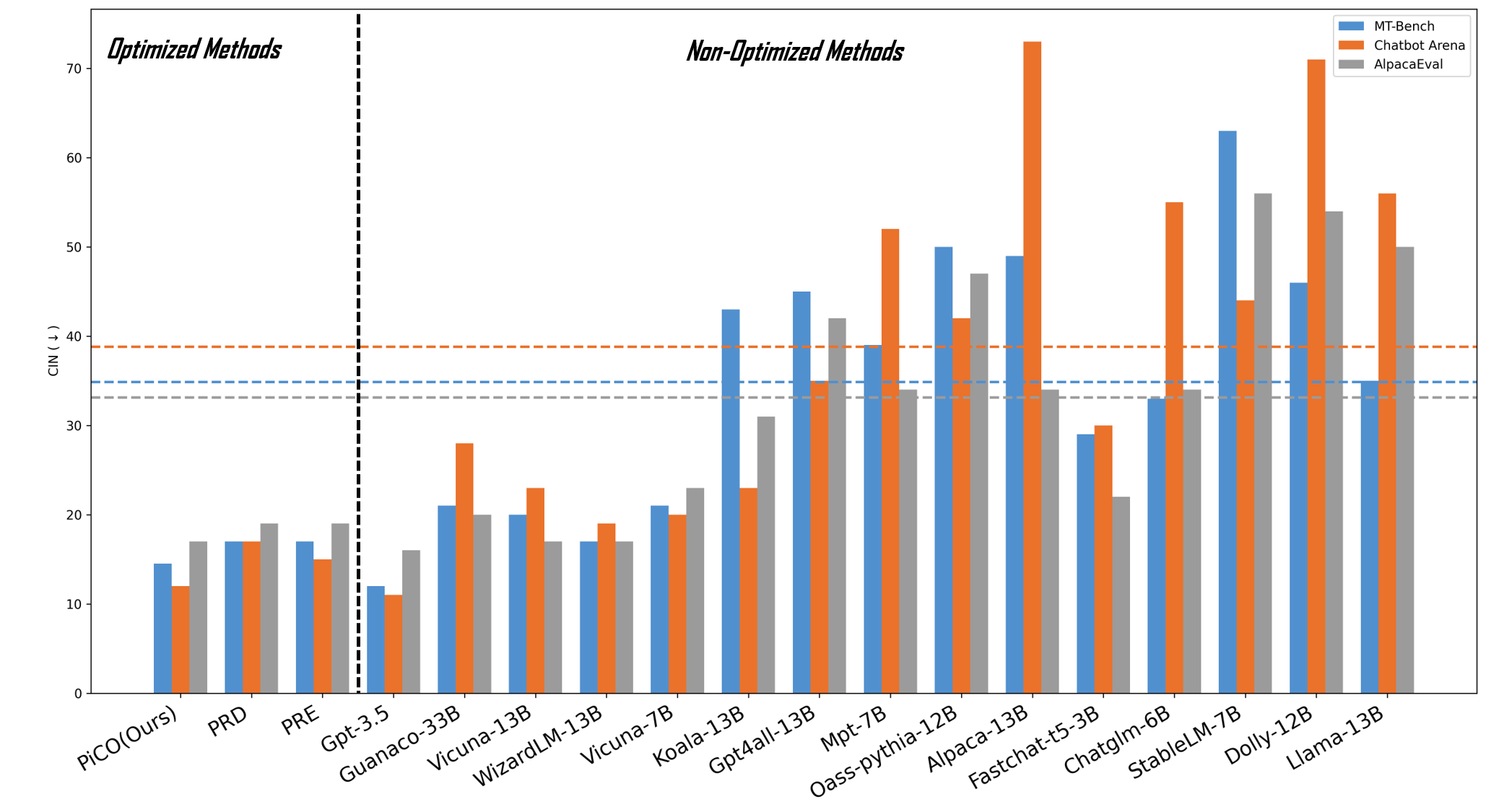}
    \caption{Comparison of performance on the CIN metric across three datasets using Unsupervised methods versus using single models, with Unsupervised methods on the left and Supervised methods on the right. The dotted line represents the average value using single models.}
    \label{fig:CIN_comparison}
\end{figure*}

\begin{figure*}[ht]
    \centering
    \includegraphics[width=1\textwidth]{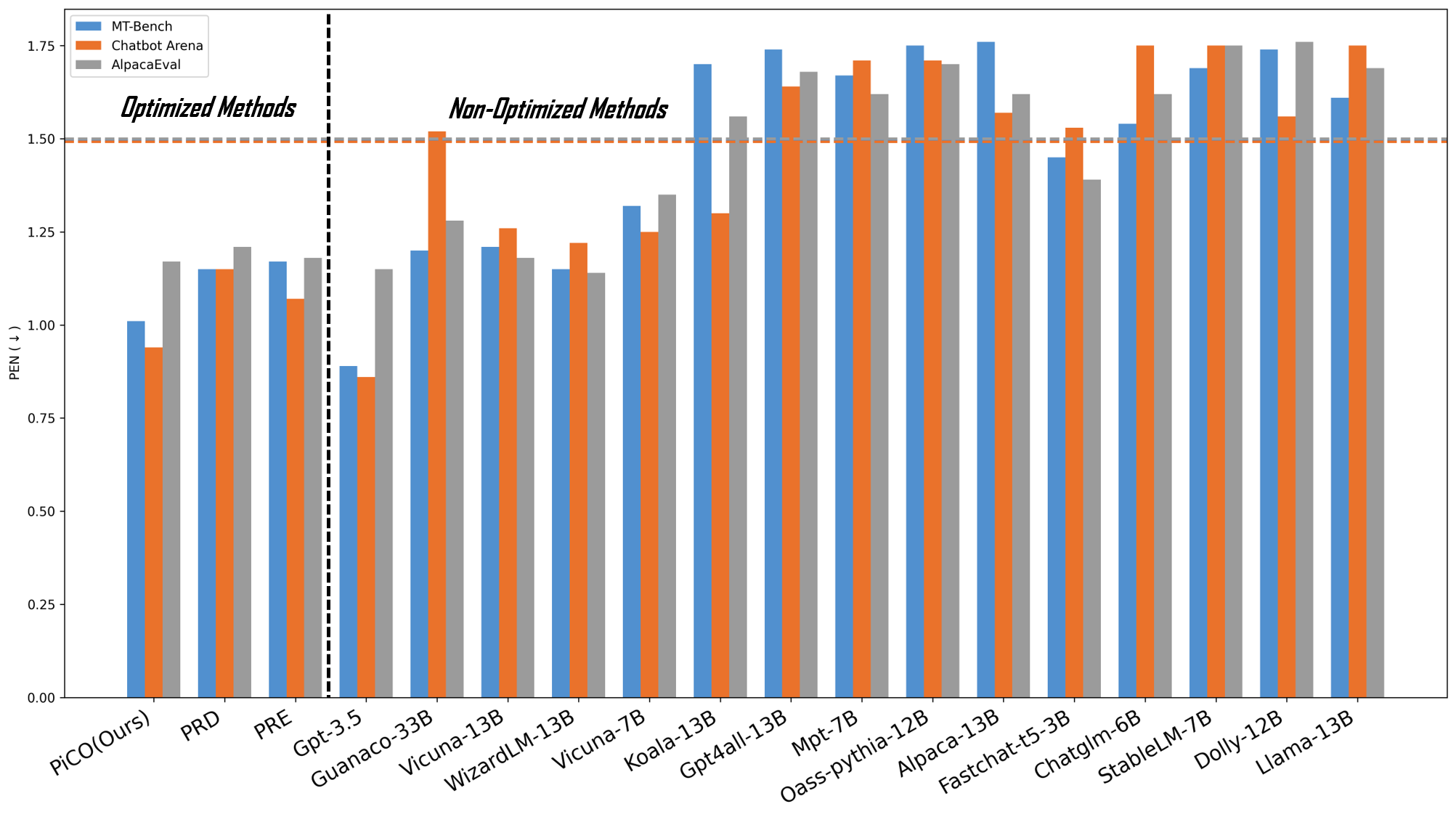}
    \caption{Comparison of performance on the PEN metric across three datasets using Unsupervised methods versus using single models, with Unsupervised methods on the left and Supervised methods on the right. The dotted line represents the average value using single models.}
    \label{fig:PEN_comparison}
\end{figure*}

\begin{figure*}[ht]
    \centering
    \includegraphics[width=1\textwidth]{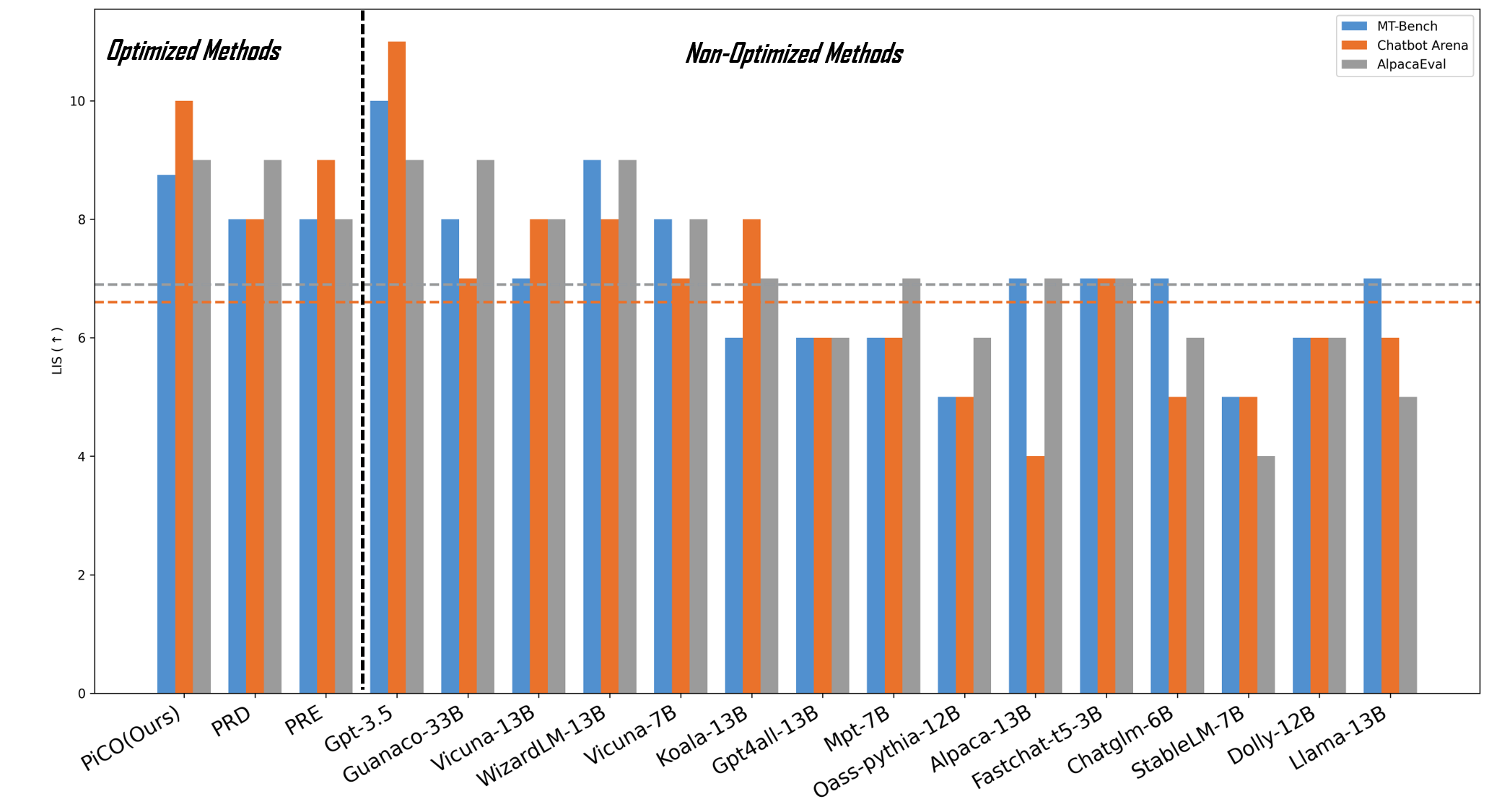}
    \caption{Comparison of performance on the LIS metric across three datasets using Unsupervised methods versus using single models, with Unsupervised methods on the left and Supervised methods on the right. The dotted line represents the average value using single models.}
    \label{fig:LIS_comparison}
\end{figure*}

In Section 3.5, we validate the effectiveness of the \textit{consistency assumption} and compare it with the Average Performance of the Reviewer Queue, i.e., employing a single LLM as the 'reviewer' to evaluate all response pairs and then calculating the average results of all LLMs. The comprehensive results compared with the Reviewer Queue are illustrated in Table\ref{table: append_upper}, Figure \ref{fig:CIN_comparison}, \ref{fig:PEN_comparison} and \ref{fig:LIS_comparison}, revealing that in the full Reviewer Queue, the performance of the vast majority of LLMs is very poor, indicating that the evaluations from most LLMs are noise. However, our 'PiCO' approach nearly matches the evaluative prowess of the pool's most capable LLM, GPT-3.5. Remarkably, given its unsupervised nature, the 'PiCO' method demonstrates the capability to mitigate the influence of noise, reaching the evaluation upper bound (the strongest LLM) within any given unknown LLM pool $M$, even in the absence of prior ranking information.

\begin{table*}[!t]
    \renewcommand{\arraystretch}{1}
    \setlength{\tabcolsep}{1pt}
    \centering
    \caption{Comparison of performance across three datasets using Unsupervised methods versus using single models in  reviewer queue.}
    \label{table: append_upper}
    \resizebox{\textwidth}{!}{\begin{tabular}{c | c c c | c c c| c c c}
    \toprule
        \multirow{2}{*}{Methods} & \multicolumn{3}{c}{MT-Bench} & \multicolumn{3}{c}{Chatbot Arena} & \multicolumn{3}{c}{AlpacaEval}  \\ 
        ~ & PEN $(\downarrow)$ & CIN$(\downarrow)$ & LIS$(\uparrow)$ & PEN $(\downarrow)$ & CIN$(\downarrow)$ & LIS$(\uparrow)$ & PEN $(\downarrow)$ & CIN$(\downarrow)$ & LIS$(\uparrow)$  \\ 
        \hline
        Gpt-3.5 & $\bm{0.97}$ & $\bm{12.00}$ & $\bm{10.00}$ & $\bm{0.85}$ & $\bm{11.00}$ & $\bm{11.00}$ & $\bm{1.15}$ & $\bm{16.00}$ & $\bm{9.00}$  \\ 
        Guanaco-33B & $1.25$ & $21.00$ & $8.00$ & $1.50$ & $28.00$ & $7.00$ & $1.26$ & $20.00$ & $9.00$  \\
        Vicuna-13B & $1.31$ & $20.00$ & $7.00$ & $1.27$ & $23.00$ & $8.00$ & $1.20$ & $17.00$ & $8.00$  \\ 
        WizardLM-13B & $1.15$ & $17.00$ & $9.00$ & $1.27$ & $19.00$ & $8.00$ & $1.17$ & $17.00$ & $9.00$  \\ 
        Vicuna-7B & $1.27$ & $21.00$ & $8.00$ & $1.30$ & $20.00$ & $7.00$ & $1.34$ & $23.00$ & $8.00$  \\ 
        Koala-13B & $1.67$ & $43.00$ & $6.00$ & $1.34$ & $23.00$ & $8.00$ & $1.54$ & $31.00$ & $7.00$  \\ 
        gpt4all-13B & $1.74$ & $45.00$ & $6.00$ & $1.60$ & $35.00$ & $6.00$ & $1.73$ & $42.00$ & $6.00$  \\ 
        Mpt-7B & $1.67$ & $39.00$ & $6.00$ & $1.72$ & $52.00$ & $6.00$ & $1.63$ & $34.00$ & $7.00$  \\ 
        Oass-pythia-12B & $1.77$ & $50.00$ & $5.00$ & $1.74$ & $42.00$ & $5.00$ & $1.70$ & $47.00$ & $6.00$  \\ 
        Alpaca-13B & $1.77$ & $49.00$ & $7.00$ & $1.60$ & $73.00$ & $4.00$ & $1.63$ & $34.00$ & $7.00$  \\ 
        FastChat-T5-3B & $1.45$ & $29.00$ & $7.00$ & $1.53$ & $30.00$ & $7.00$ & $1.30$ & $22.00$ & $7.00$  \\ 
        ChatGLM-6B & $1.59$ & $33.00$ & $7.00$ & $1.71$ & $55.00$ & $5.00$ & $1.63$ & $34.00$ & $6.00$  \\ 
        StableLM-7B & $1.68$ & $63.00$ & $5.00$ & $1.75$ & $44.00$ & $5.00$ & $1.72$ & $56.00$ & $4.00$  \\ 
        Dolly-12B & $1.76$ & $46.00$ & $6.00$ & $1.57$ & $71.00$ & $6.00$ & $1.75$ & $54.00$ & $6.00$  \\ 
        LLaMA-13B & $1.60$ & $35.00$ & $7.00$ & $1.76$ & $56.00$ & $6.00$ & $1.70$ & $50.00$ & $5.00$ \\ \hline
        Average Performance of All Review LLMs & $1.51$ & $34.87$ & $6.93$ & $1.50$ & $38.80$ & $6.60$ & $1.50$ & $33.13$ & $6.93$  \\ \hline
        PRD\citep{li2023prd} & $1.15$ & $17.00$ & $8.00$ & $1.15$ & $17.00$ & $8.00$ & $1.21$ & $19.00$ & $\underline{9.00}$  \\ 
        PRE\citep{chu2024pre} & $1.17$ & $17.00$ & $8.00$ & $1.07$ & $15.00$ & $9.00$ & $1.18$ & $19.00$ & $8.00$  \\ 
        PiCO (Ours) & $\underline{1.01}$ & $\underline{14.50}$ & $\underline{8.75}$ & $\underline{0.94}$ & $\underline{12.00}$ & $\underline{10.00}$ & $\underline{1.17}$ & $\underline{17.00}$ & $\underline{9.00}$  \\
        \toprule
    \end{tabular}}
\end{table*}

\section{Selected Models and Optimized Ranking}
\label{appendix:F}
For our analysis, we meticulously selected 15 LLMs spanning a variety of architectures, encompassing both open-source and closed-source models, as detailed in the subsequent table. Our curated selection features prominent LLMs including the closed-source "gpt-3.5-turbo," "chatglm" which is predicated on the encoder-decoder framework, "fastchat-t5-3b" that leverages Google's T5 (Text-to-Text Transfer Transformer) architecture, and "llama-13b" founded on the GPT architectural principles.

We have comprehensively detailed the ranking outcomes across three distinct datasets for our comparative analysis, incorporating the optimized model rankings, names, and their respective scores. As delineated in Appendix \ref{appendix:C}, the PiCO (Ours) is capable of employing various scoring mechanisms, thereby facilitating the presentation of ranking outcomes on three datasets utilizing both the Elo and Rank mechanisms. Furthermore, we have also enumerated the ranking results for PRD and PRE methodologies across the three datasets, offering a holistic view of the competitive landscape.

\newpage
\subsection{PiCO}

\begin{tcolorbox}[colback=gray!20!white, colbacktitle=white, coltitle=black, colframe=black!75!black, boxrule=0.7pt, halign title=center, title=\textbf{Grade-Elo-Chatbot}]
$\#$\textcolor{lightblue}{1  \textbf{Gpt-3.5}} | Grade: \textcolor{lightblue}{9205.162109375}\\
$\#$\textcolor{lightblue}{2  \textbf{WizardLM-13B}} | Grade: \textcolor{lightblue}{9143.46875}\\
$\#$\textcolor{lightblue}{3  \textbf{Guanaco-33B}} | Grade: \textcolor{lightblue}{5886.92626953125}\\
$\#$\textcolor{lightblue}{4  \textbf{Vicuna-7B}} | Grade: \textcolor{lightblue}{5368.9462890625}\\
$\#$\textcolor{lightblue}{5  \textbf{Vicuna-13B}} | Grade: \textcolor{lightblue}{5216.79541015625}\\
$\#$\textcolor{lightblue}{6  \textbf{Koala-13B}} | Grade: \textcolor{lightblue}{3545.1171875} | Eliminated\\
$\#$\textcolor{lightblue}{7  \textbf{Mpt-7B}} | Grade: \textcolor{lightblue}{962.99462890625} | Eliminated\\
$\#$\textcolor{lightblue}{8  \textbf{Gpt4all-13B}} | Grade: \textcolor{lightblue}{652.4602661132812} | Eliminated\\
$\#$\textcolor{lightblue}{9  \textbf{Chatglm-6B}} | Grade: \textcolor{lightblue}{417.1375427246094} | Eliminated\\
$\#$\textcolor{lightblue}{10  \textbf{Oasst-pythia-12B}} | Grade: \textcolor{lightblue}{-898.2676391601562} | Eliminated\\
$\#$\textcolor{lightblue}{11  \textbf{Fastchat-t5-3B}} | Grade: \textcolor{lightblue}{-1251.7183837890625} | Eliminated\\
$\#$\textcolor{lightblue}{12  \textbf{StableLM-7B}} | Grade: \textcolor{lightblue}{-2232.66943359375}  | Eliminated\\
$\#$\textcolor{lightblue}{13  \textbf{Dolly-12B}} | Grade: \textcolor{lightblue}{-3163.540283203125} | Eliminated\\
$\#$\textcolor{lightblue}{14  \textbf{Llama-13B}} | Grade: \textcolor{lightblue}{-3648.37841796875} | Eliminated\\
$\#$\textcolor{lightblue}{15  \textbf{Alpaca-13B}} | Grade: \textcolor{lightblue}{-14204.3984375} | Eliminated
\vspace*{-0.5\baselineskip}
\end{tcolorbox}

\begin{tcolorbox}[colback=gray!20!white, colbacktitle=white, coltitle=black, colframe=black!75!black, boxrule=0.7pt, halign title=center, title=\textbf{Grade-Elo-AlpacaEval}]
$\#$\textcolor{lightblue}{1  \textbf{WizardLM-13B}} | Grade: \textcolor{lightblue}{8662.7158203125}\\
$\#$\textcolor{lightblue}{2  \textbf{Vicuna-13B}} | Grade: \textcolor{lightblue}{5586.46630859375}\\
$\#$\textcolor{lightblue}{3  \textbf{Guanaco-33B}} | Grade: \textcolor{lightblue}{5445.341796875}\\
$\#$\textcolor{lightblue}{4  \textbf{Vicuna-7B}} | Grade: \textcolor{lightblue}{5374.2314453125}\\
$\#$\textcolor{lightblue}{5  \textbf{Gpt-3.5}} | Grade: \textcolor{lightblue}{4845.91552734375}\\
$\#$\textcolor{lightblue}{6  \textbf{Koala-13B}} | Grade: \textcolor{lightblue}{4338.77783203125} | Eliminated\\
$\#$\textcolor{lightblue}{7  \textbf{Chatglm-6B}} | Grade: \textcolor{lightblue}{2293.4208984375} | Eliminated\\
$\#$\textcolor{lightblue}{8  \textbf{Gpt4all-13B}} | Grade: \textcolor{lightblue}{2080.511962890625} | Eliminated\\
$\#$\textcolor{lightblue}{9  \textbf{Mpt-7B}} | Grade: \textcolor{lightblue}{1694.4945068359375} | Eliminated\\
$\#$\textcolor{lightblue}{10  \textbf{Fastchat-t5-3B}} | Grade: \textcolor{lightblue}{1371.94287109375} | Eliminated\\
$\#$\textcolor{lightblue}{11  \textbf{Oasst-pythia-12B}} | Grade: \textcolor{lightblue}{-665.8685302734375} | Eliminated\\
$\#$\textcolor{lightblue}{12  \textbf{StableLM-7B}} | Grade: \textcolor{lightblue}{-1343.5838623046875} | Eliminated\\
$\#$\textcolor{lightblue}{13  \textbf{Dolly-12B}} | Grade: \textcolor{lightblue}{-5377.13427734375} | Eliminated\\
$\#$\textcolor{lightblue}{14  \textbf{Llama-13B}} | Grade: \textcolor{lightblue}{-5847.59130859375} | Eliminated\\
$\#$\textcolor{lightblue}{15  \textbf{Alpaca-13B}} | Grade: \textcolor{lightblue}{-13459.6162109375} | Eliminated
\end{tcolorbox}

\begin{tcolorbox}[colback=gray!20!white, colbacktitle=white, coltitle=black, colframe=black!75!black, boxrule=0.7pt, halign title=center, title=\textbf{Grade-Elo-MT$\_$Bench}]
$\#$\textcolor{lightblue}{1  \textbf{WizardLM-13B}} | Grade: \textcolor{lightblue}{2178.10302734375}\\
$\#$\textcolor{lightblue}{2  \textbf{Vicuna-13B}} | Grade: \textcolor{lightblue}{1720.1114501953125}\\
$\#$\textcolor{lightblue}{3  \textbf{Guanaco-33B}} | Grade: \textcolor{lightblue}{1704.1832275390625}\\
$\#$\textcolor{lightblue}{4  \textbf{Vicuna-7B}} | Grade: \textcolor{lightblue}{1659.2799072265625}\\
$\#$\textcolor{lightblue}{5  \textbf{Gpt-3.5}} | Grade: \textcolor{lightblue}{1535.8819580078125}\\
$\#$\textcolor{lightblue}{6  \textbf{Mpt-7B}} | Grade: \textcolor{lightblue}{1338.5235595703125} | Eliminated\\
$\#$\textcolor{lightblue}{7  \textbf{Koala-13B}} | Grade: \textcolor{lightblue}{1267.9747314453125} | Eliminated\\
$\#$\textcolor{lightblue}{8  \textbf{Chatglm-6B}} | Grade: \textcolor{lightblue}{1011.7701416015625} | Eliminated\\
$\#$\textcolor{lightblue}{9  \textbf{Gpt4all-13B}} | Grade: \textcolor{lightblue}{976.5963745117188} | Eliminated\\
$\#$\textcolor{lightblue}{10  \textbf{Oasst-pythia-12B}} | Grade: \textcolor{lightblue}{779.3573608398438} | Eliminated\\
$\#$\textcolor{lightblue}{11  \textbf{StableLM-7B}} | Grade: \textcolor{lightblue}{512.1678466796875} | Eliminated\\
$\#$\textcolor{lightblue}{12  \textbf{Alpaca-13B}} | Grade: \textcolor{lightblue}{334.9879455566406} | Eliminated\\
$\#$\textcolor{lightblue}{13  \textbf{Fastchat-t5-3B}} | Grade: \textcolor{lightblue}{303.5980529785156} | Eliminated\\
$\#$\textcolor{lightblue}{14  \textbf{Dolly-12B}} | Grade: \textcolor{lightblue}{72.63818359375} | Eliminated\\
$\#$\textcolor{lightblue}{15  \textbf{Llama-13B}} | Grade: \textcolor{lightblue}{-395.19921875} | Eliminated
\end{tcolorbox}

\begin{tcolorbox}[colback=gray!20!white, colbacktitle=white, coltitle=black, colframe=black!75!black, boxrule=0.7pt, halign title=center, title=\textbf{Grade-Rank-Chatbot}]
$\#$\textcolor{lightblue}{1  \textbf{WizardLM-13B}} | Grade: \textcolor{lightblue}{0.30809280276298523}\\
$\#$\textcolor{lightblue}{2  \textbf{Gpt-3.5}} | Grade: \textcolor{lightblue}{0.293962299823761}\\
$\#$\textcolor{lightblue}{3  \textbf{Guanaco-33B}} | Grade: \textcolor{lightblue}{0.28587597608566284}\\
$\#$\textcolor{lightblue}{4  \textbf{Vicuna-7B}} | Grade: \textcolor{lightblue}{0.28212910890579224}\\
$\#$\textcolor{lightblue}{5  \textbf{Vicuna-13B}} | Grade: \textcolor{lightblue}{0.27900218963623047}\\
$\#$\textcolor{lightblue}{6  \textbf{Koala-13B}} | Grade: \textcolor{lightblue}{0.2672431766986847} | Eliminated\\
$\#$\textcolor{lightblue}{7  \textbf{Mpt-7B}} | Grade: \textcolor{lightblue}{0.2500302195549011} | Eliminated\\
$\#$\textcolor{lightblue}{8  \textbf{Gpt4all-13B}} | Grade: \textcolor{lightblue}{0.24746862053871155} | Eliminated\\
$\#$\textcolor{lightblue}{9  \textbf{Chatglm-6B}} | Grade: \textcolor{lightblue}{0.2466953843832016} | Eliminated\\
$\#$\textcolor{lightblue}{10  \textbf{Oasst-pythia-12B}} | Grade: \textcolor{lightblue}{0.23637069761753082} | Eliminated\\
$\#$\textcolor{lightblue}{11  \textbf{Fastchat-t5-3B}} | Grade: \textcolor{lightblue}{0.2350562959909439} | Eliminated\\
$\#$\textcolor{lightblue}{12  \textbf{StableLM-7B}} | Grade: \textcolor{lightblue}{0.22843806445598602} | Eliminated\\
$\#$\textcolor{lightblue}{13  \textbf{Dolly-12B}} | Grade: \textcolor{lightblue}{0.22219440340995789} | Eliminated\\
$\#$\textcolor{lightblue}{14  \textbf{Llama-13B}} | Grade: \textcolor{lightblue}{0.2165679931640625} | Eliminated\\
$\#$\textcolor{lightblue}{15  \textbf{Alpaca-13B}} | Grade: \textcolor{lightblue}{0.13975904881954193} | Eliminated
\end{tcolorbox}

\begin{tcolorbox}[colback=gray!20!white, colbacktitle=white, coltitle=black, colframe=black!75!black, boxrule=0.7pt, halign title=center, title=\textbf{Grade-Rank-AlpacaEval}]
$\#$\textcolor{lightblue}{1  \textbf{WizardLM-13B}} | Grade: \textcolor{lightblue}{0.4019235074520111}\\
$\#$\textcolor{lightblue}{2  \textbf{Vicuna-13B}} | Grade: \textcolor{lightblue}{0.36745429039001465}\\
$\#$\textcolor{lightblue}{3  \textbf{Guanaco-33B}} | Grade: \textcolor{lightblue}{0.3664878010749817}\\
$\#$\textcolor{lightblue}{4  \textbf{Vicuna-7B}} | Grade: \textcolor{lightblue}{0.36541733145713806}\\
$\#$\textcolor{lightblue}{5  \textbf{Gpt-3.5}} | Grade: \textcolor{lightblue}{0.36000365018844604}\\
$\#$\textcolor{lightblue}{6  \textbf{Koala-13B}} | Grade: \textcolor{lightblue}{0.3544933795928955} | Eliminated\\
$\#$\textcolor{lightblue}{7  \textbf{Chatglm-6B}} | Grade: \textcolor{lightblue}{0.3319571018218994} | Eliminated\\
$\#$\textcolor{lightblue}{8  \textbf{Gpt4all-13B}} | Grade: \textcolor{lightblue}{0.3306528627872467} | Eliminated\\
$\#$\textcolor{lightblue}{9  \textbf{Mpt-7B}} | Grade: \textcolor{lightblue}{0.32641729712486267} | Eliminated\\
$\#$\textcolor{lightblue}{10  \textbf{Fastchat-t5-3B}} | Grade: \textcolor{lightblue}{0.32173293828964233} | Eliminated\\
$\#$\textcolor{lightblue}{11  \textbf{Oasst-pythia-12B}} | Grade: \textcolor{lightblue}{0.2999681532382965} | Eliminated\\
$\#$\textcolor{lightblue}{12  \textbf{StableLM-7B}} | Grade: \textcolor{lightblue}{0.2932431995868683} | Eliminated\\
$\#$\textcolor{lightblue}{13  \textbf{Dolly-12B}} | Grade: \textcolor{lightblue}{0.24777530133724213} | Eliminated\\
$\#$\textcolor{lightblue}{14  \textbf{Llama-13B}} | Grade: \textcolor{lightblue}{0.24381506443023682} | Eliminated\\
$\#$\textcolor{lightblue}{15  \textbf{Alpaca-13B}} | Grade: \textcolor{lightblue}{0.16114839911460876}
\end{tcolorbox}

\begin{tcolorbox}[colback=gray!20!white, colbacktitle=white, coltitle=black, colframe=black!75!black, boxrule=0.7pt, halign title=center, title=\textbf{Grade-Rank-MT$\_$Bench}]
$\#$\textcolor{lightblue}{1  \textbf{WizardLM-13B}} | Grade: \textcolor{lightblue}{0.2994651198387146}\\
$\#$\textcolor{lightblue}{2  \textbf{Vicuna-13B}} | Grade: \textcolor{lightblue}{0.2809261679649353}\\
$\#$\textcolor{lightblue}{3  \textbf{Guanaco-33B}} | Grade: \textcolor{lightblue}{0.2767307460308075}\\
$\#$\textcolor{lightblue}{4  \textbf{Vicuna-7B}} | Grade: \textcolor{lightblue}{0.2758147716522217}\\
$\#$\textcolor{lightblue}{5  \textbf{Gpt-3.5}} | Grade: \textcolor{lightblue}{0.27261608839035034}\\
$\#$\textcolor{lightblue}{6  \textbf{Mpt-7B}} | Grade: \textcolor{lightblue}{0.26338690519332886} | Eliminated\\
$\#$\textcolor{lightblue}{7  \textbf{Koala-13B}} | Grade: \textcolor{lightblue}{0.2613368630409241} | Eliminated\\
$\#$\textcolor{lightblue}{8  \textbf{Gpt4all-13B}} | Grade: \textcolor{lightblue}{0.24908888339996338} | Eliminated\\
$\#$\textcolor{lightblue}{9  \textbf{Chatglm-6B}} | Grade: \textcolor{lightblue}{0.24898234009742737} | Eliminated\\
$\#$\textcolor{lightblue}{10  \textbf{Oasst-pythia-12B}} | Grade: \textcolor{lightblue}{0.2415400892496109} | Eliminated\\
$\#$\textcolor{lightblue}{11  \textbf{StableLM-7B}} | Grade: \textcolor{lightblue}{0.2299075722694397} | Eliminated\\
$\#$\textcolor{lightblue}{12  \textbf{Alpaca-13B}} | Grade: \textcolor{lightblue}{0.22171474993228912} | Eliminated\\
$\#$\textcolor{lightblue}{13  \textbf{Fastchat-t5-3B}} | Grade: \textcolor{lightblue}{0.221677765250206} | Eliminated\\
$\#$\textcolor{lightblue}{14  \textbf{Dolly-12B}} | Grade: \textcolor{lightblue}{0.21185410022735596} | Eliminated\\
$\#$\textcolor{lightblue}{15  \textbf{Llama-13B}} | Grade: \textcolor{lightblue}{0.192665234208107} | Eliminated
\end{tcolorbox}

\subsection{PRD}

\begin{tcolorbox}[colback=gray!20!white, colbacktitle=white, coltitle=black, colframe=black!75!black, boxrule=0.7pt, halign title=center, title=\textbf{PRD-Chatbot}]
$\#$\textcolor{lightblue}{1  \textbf{WizardLM-13B}} | Grade: \textcolor{lightblue}{5565.28271484375}\\
$\#$\textcolor{lightblue}{2  \textbf{Gpt-3.5}} | Grade: \textcolor{lightblue}{4613.22900390625}\\
$\#$\textcolor{lightblue}{3  \textbf{Guanaco-33B}} | Grade: \textcolor{lightblue}{3423.588134765625}\\
$\#$\textcolor{lightblue}{4  \textbf{Vicuna-7B}} | Grade: \textcolor{lightblue}{2985.4892578125}\\
$\#$\textcolor{lightblue}{5  \textbf{Vicuna-13B}} | Grade: \textcolor{lightblue}{2972.15673828125}\\
$\#$\textcolor{lightblue}{6  \textbf{Koala-13B}} | Grade: \textcolor{lightblue}{2237.70751953125}\\
$\#$\textcolor{lightblue}{7  \textbf{Chatglm-6B}} | Grade: \textcolor{lightblue}{875.373779296875}\\
$\#$\textcolor{lightblue}{8  \textbf{Mpt-7B}} | Grade: \textcolor{lightblue}{602.46923828125}\\
$\#$\textcolor{lightblue}{9  \textbf{Gpt4all-13B}} | Grade: \textcolor{lightblue}{356.06243896484375}\\
$\#$\textcolor{lightblue}{10  \textbf{Fastchat-t5-3B}} | Grade: \textcolor{lightblue}{184.89663696289062}\\
$\#$\textcolor{lightblue}{11  \textbf{Dolly-12B}} | Grade: \textcolor{lightblue}{52.10746765136719}\\
$\#$\textcolor{lightblue}{12  \textbf{Oasst-pythia-12B}} | Grade: \textcolor{lightblue}{-307.49908447265625}\\
$\#$\textcolor{lightblue}{13  \textbf{StableLM-7B}} | Grade: \textcolor{lightblue}{-691.4453735351562}\\
$\#$\textcolor{lightblue}{14  \textbf{Llama-13B}} | Grade: \textcolor{lightblue}{-848.1654052734375}\\
$\#$\textcolor{lightblue}{15  \textbf{Alpaca-13B}} | Grade: \textcolor{lightblue}{-7020.923828125}
\end{tcolorbox}

\begin{tcolorbox}[colback=gray!20!white, colbacktitle=white, coltitle=black, colframe=black!75!black, boxrule=0.7pt, halign title=center, title=\textbf{PRD-AlpacaEval}]
$\#$\textcolor{lightblue}{1  \textbf{WizardLM-13B}} | Grade: \textcolor{lightblue}{5469.75634765625}\\
$\#$\textcolor{lightblue}{2  \textbf{Guanaco-33B}} | Grade: \textcolor{lightblue}{3707.014892578125}\\
$\#$\textcolor{lightblue}{3  \textbf{Vicuna-13B}} | Grade: \textcolor{lightblue}{3618.63427734375}\\
$\#$\textcolor{lightblue}{4  \textbf{Vicuna-7B}} | Grade: \textcolor{lightblue}{3569.389892578125}\\
$\#$\textcolor{lightblue}{5  \textbf{Gpt-3.5}} | Grade: \textcolor{lightblue}{3197.755615234375}\\
$\#$\textcolor{lightblue}{6  \textbf{Koala-13B}} | Grade: \textcolor{lightblue}{2893.642578125}\\
$\#$\textcolor{lightblue}{7  \textbf{Chatglm-6B}} | Grade: \textcolor{lightblue}{1847.1300048828125}\\
$\#$\textcolor{lightblue}{8  \textbf{Fastchat-t5-3B}} | Grade: \textcolor{lightblue}{1585.66943359375}\\
$\#$\textcolor{lightblue}{9  \textbf{Gpt4all-13B}} | Grade: \textcolor{lightblue}{1561.145751953125}\\
$\#$\textcolor{lightblue}{10  \textbf{Mpt-7B}} | Grade: \textcolor{lightblue}{1332.3753662109375}\\
$\#$\textcolor{lightblue}{11  \textbf{StableLM-7B}} | Grade: \textcolor{lightblue}{-33.00855255126953}\\
$\#$\textcolor{lightblue}{12  \textbf{Oasst-pythia-12B}} | Grade: \textcolor{lightblue}{-92.68387603759766}\\
$\#$\textcolor{lightblue}{13  \textbf{Dolly-12B}} | Grade: \textcolor{lightblue}{-3013.588623046875}\\
$\#$\textcolor{lightblue}{14  \textbf{Llama-13B}} | Grade: \textcolor{lightblue}{-3211.0302734375}\\
$\#$\textcolor{lightblue}{15  \textbf{Alpaca-13B}} | Grade: \textcolor{lightblue}{-7432.3701171875}
\end{tcolorbox}

\begin{tcolorbox}[colback=gray!20!white, colbacktitle=white, coltitle=black, colframe=black!75!black, boxrule=0.7pt, halign title=center, title=\textbf{PRD-MT$\_$Bench}]
$\#$\textcolor{lightblue}{1  \textbf{WizardLM-13B}} | Grade: \textcolor{lightblue}{1811.64697265625}\\
$\#$\textcolor{lightblue}{2  \textbf{Vicuna-13B}} | Grade: \textcolor{lightblue}{1537.8084716796875}\\
$\#$\textcolor{lightblue}{3  \textbf{Guanaco-33B}} | Grade: \textcolor{lightblue}{1481.1739501953125}\\
$\#$\textcolor{lightblue}{4  \textbf{Vicuna-7B}} | Grade: \textcolor{lightblue}{1401.5194091796875}\\
$\#$\textcolor{lightblue}{5  \textbf{Gpt-3.5}} | Grade: \textcolor{lightblue}{1272.8072509765625}\\
$\#$\textcolor{lightblue}{6  \textbf{Mpt-7B}} | Grade: \textcolor{lightblue}{1186.5518798828125}\\
$\#$\textcolor{lightblue}{7  \textbf{Chatglm-6B}} | Grade: \textcolor{lightblue}{1166.6246337890625}\\
$\#$\textcolor{lightblue}{8  \textbf{Koala-13B}} | Grade: \textcolor{lightblue}{1124.2513427734375}\\
$\#$\textcolor{lightblue}{9  \textbf{Gpt4all-13B}} | Grade: \textcolor{lightblue}{871.2874755859375}\\
$\#$\textcolor{lightblue}{10  \textbf{Oasst-pythia-12B}} | Grade: \textcolor{lightblue}{855.3653564453125}\\
$\#$\textcolor{lightblue}{11  \textbf{StableLM-7B}} | Grade: \textcolor{lightblue}{782.702880859375}\\
$\#$\textcolor{lightblue}{12  \textbf{Fastchat-t5-3B}} | Grade: \textcolor{lightblue}{636.966064453125}\\
$\#$\textcolor{lightblue}{13  \textbf{Alpaca-13B}} | Grade: \textcolor{lightblue}{414.9374694824219}\\
$\#$\textcolor{lightblue}{14  \textbf{Dolly-12B}} | Grade: \textcolor{lightblue}{377.5018005371094}\\
$\#$\textcolor{lightblue}{15  \textbf{Llama-13B}} | Grade: \textcolor{lightblue}{78.90127563476562}
\end{tcolorbox}

\subsection{PRE}
\begin{tcolorbox}[colback=gray!20!white, colbacktitle=white, coltitle=black, colframe=black!75!black, boxrule=0.7pt, halign title=center, title=\textbf{PRE-Chatbot}]
$\#$\textcolor{lightblue}{1  \textbf{WizardLM-13B}} | Grade: \textcolor{lightblue}{1113.7034715479742}\\
$\#$\textcolor{lightblue}{2  \textbf{Gpt-3.5}} | Grade: \textcolor{lightblue}{1076.1116664199608}\\
$\#$\textcolor{lightblue}{3  \textbf{Guanaco-33B}} | Grade: \textcolor{lightblue}{1067.441581415147}\\
$\#$\textcolor{lightblue}{4  \textbf{Vicuna-13B}} | Grade: \textcolor{lightblue}{1057.702184441485}\\
$\#$\textcolor{lightblue}{5  \textbf{Vicuna-7B}} | Grade: \textcolor{lightblue}{1043.4840340151043}\\
$\#$\textcolor{lightblue}{6  \textbf{Koala-13B}} | Grade: \textcolor{lightblue}{1030.4455842017508} | Eliminated\\
$\#$\textcolor{lightblue}{7  \textbf{Chatglm-6B}} | Grade: \textcolor{lightblue}{1012.4487557424748} | Eliminated\\
$\#$\textcolor{lightblue}{8  \textbf{Mpt-7B}} | Grade: \textcolor{lightblue}{1000.487230109001} | Eliminated\\
$\#$\textcolor{lightblue}{9  \textbf{Gpt4all-13B}} | Grade: \textcolor{lightblue}{1000.4111397038492} | Eliminated\\
$\#$\textcolor{lightblue}{10  \textbf{Fastchat-t5-3B}} | Grade: \textcolor{lightblue}{992.3732179832363} | Eliminated\\
$\#$\textcolor{lightblue}{11  \textbf{Oasst-pythia-12B}} | Grade: \textcolor{lightblue}{977.5217305871272} | Eliminated\\
$\#$\textcolor{lightblue}{12  \textbf{StableLM-7B}} | Grade: \textcolor{lightblue}{970.3665926795535} | Eliminated\\
$\#$\textcolor{lightblue}{13  \textbf{Llama-13B}} | Grade: \textcolor{lightblue}{929.6268868888149} | Eliminated\\
$\#$\textcolor{lightblue}{14  \textbf{Dolly-12B}} | Grade: \textcolor{lightblue}{929.1943463130976} | Eliminated\\
$\#$\textcolor{lightblue}{15  \textbf{Alpaca-13B}} | Grade: \textcolor{lightblue}{798.6815779514078} | Eliminated
\end{tcolorbox}

\begin{tcolorbox}[colback=gray!20!white, colbacktitle=white, coltitle=black, colframe=black!75!black, boxrule=0.7pt, halign title=center, title=\textbf{PRE-AlpacaEval}]
$\#$\textcolor{lightblue}{1  \textbf{WizardLM-13B}} | Grade: \textcolor{lightblue}{1127.822808841937}\\
$\#$\textcolor{lightblue}{2  \textbf{Vicuna-7B}} | Grade: \textcolor{lightblue}{1077.1823389450524}\\
$\#$\textcolor{lightblue}{3  \textbf{Vicuna-13B}} | Grade: \textcolor{lightblue}{1075.4338443616266}\\
$\#$\textcolor{lightblue}{4  \textbf{Guanaco-33B}} | Grade: \textcolor{lightblue}{1074.8043135229418}\\
$\#$\textcolor{lightblue}{5  \textbf{Gpt-3.5}} | Grade: \textcolor{lightblue}{1065.305736105376}\\
$\#$\textcolor{lightblue}{6  \textbf{Gpt4all-13B}} | Grade: \textcolor{lightblue}{1039.4091630861865} | Eliminated\\
$\#$\textcolor{lightblue}{7  \textbf{Koala-13B}} | Grade: \textcolor{lightblue}{1038.205749976473} | Eliminated\\
$\#$\textcolor{lightblue}{8  \textbf{Mpt-7B}} | Grade: \textcolor{lightblue}{1032.2893401162178} | Eliminated\\
$\#$\textcolor{lightblue}{9  \textbf{Chatglm-6B}} | Grade: \textcolor{lightblue}{1027.1937496918501} | Eliminated\\
$\#$\textcolor{lightblue}{10  \textbf{Fastchat-t5-3B}} | Grade: \textcolor{lightblue}{992.3481168791307} | Eliminated\\
$\#$\textcolor{lightblue}{11  \textbf{StableLM-7B}} | Grade: \textcolor{lightblue}{979.3894141445692} | Eliminated\\
$\#$\textcolor{lightblue}{12  \textbf{Oasst-pythia-12B}} | Grade: \textcolor{lightblue}{940.6438439723215} | Eliminated\\
$\#$\textcolor{lightblue}{13  \textbf{Dolly-12B}} | Grade: \textcolor{lightblue}{886.1412110662756} | Eliminated\\
$\#$\textcolor{lightblue}{14  \textbf{Llama-13B}} | Grade: \textcolor{lightblue}{880.0797724297793} | Eliminated\\
$\#$\textcolor{lightblue}{15  \textbf{Alpaca-13B}} | Grade: \textcolor{lightblue}{763.7505968602533} | Eliminated
\end{tcolorbox}

\begin{tcolorbox}[colback=gray!20!white, colbacktitle=white, coltitle=black, colframe=black!75!black, boxrule=0.7pt, halign title=center, title=\textbf{PRE-MT$\_$Bench}]
$\#$\textcolor{lightblue}{1  \textbf{WizardLM-13B}} | Grade: \textcolor{lightblue}{1065.5843776639435}\\
$\#$\textcolor{lightblue}{2  \textbf{Vicuna-13B}} | Grade: \textcolor{lightblue}{1062.3934138040302}\\
$\#$\textcolor{lightblue}{3  \textbf{Guanaco-33B}} | Grade: \textcolor{lightblue}{1052.2206466556906}\\
$\#$\textcolor{lightblue}{4  \textbf{Vicuna-7B}} | Grade: \textcolor{lightblue}{1035.1112817247572}\\
$\#$\textcolor{lightblue}{5  \textbf{Gpt-3.5}} | Grade: \textcolor{lightblue}{1029.8316754711038}\\
$\#$\textcolor{lightblue}{6  \textbf{Koala-13B}} | Grade: \textcolor{lightblue}{1024.9307662983267} | Eliminated\\
$\#$\textcolor{lightblue}{7  \textbf{Chatglm-6B}} | Grade: \textcolor{lightblue}{1020.5238960907612} | Eliminated\\
$\#$\textcolor{lightblue}{8  \textbf{Mpt-7B}} | Grade: \textcolor{lightblue}{1014.0683255081057} | Eliminated\\
$\#$\textcolor{lightblue}{9  \textbf{Gpt4all-13B}} | Grade: \textcolor{lightblue}{991.7142639623017} | Eliminated\\
$\#$\textcolor{lightblue}{10  \textbf{StableLM-7B}} | Grade: \textcolor{lightblue}{979.8443261256327} | Eliminated\\
$\#$\textcolor{lightblue}{11  \textbf{Oasst-pythia-12B}} | Grade: \textcolor{lightblue}{977.9930430111322} | Eliminated\\
$\#$\textcolor{lightblue}{12  \textbf{Fastchat-t5-3B}} | Grade: \textcolor{lightblue}{953.0776159143571} | Eliminated\\
$\#$\textcolor{lightblue}{13  \textbf{Alpaca-13B}} | Grade: \textcolor{lightblue}{949.129770731626} | Eliminated\\
$\#$\textcolor{lightblue}{14  \textbf{Dolly-12B}} | Grade: \textcolor{lightblue}{928.511065779112} | Eliminated\\
$\#$\textcolor{lightblue}{15  \textbf{Llama-13B}} | Grade: \textcolor{lightblue}{915.0655312591185} | Eliminated
\end{tcolorbox}

\newpage

\end{document}